\documentclass[10pt,twocolumn,letterpaper]{article}
\usepackage[pagenumbers]{wacv} 

\usepackage{xcolor}
\usepackage{multirow}

\usepackage[normalem]{ulem}   
\usepackage{subcaption}       

\definecolor{wacvblue}{rgb}{0.21,0.49,0.74}
\definecolor{rebuttal}{rgb}{1.0, 0, 0}
\usepackage[pagebackref,breaklinks,colorlinks,allcolors=wacvblue]{hyperref}

\newif\ifreviewmode
\reviewmodefalse  

\definecolor{fxQT}{rgb}{0.278, 0.780, 0.255}   
\definecolor{gpqz}{rgb}{0.851, 0.373, 0.008}   
\definecolor{MRVu}{rgb}{0.459, 0.239, 0.718}   
\definecolor{shared}{rgb}{0.121, 0.470, 0.706} 

\newcommand{\Rhi}[2]{%
  \ifreviewmode\textcolor{#1}{#2}\else#2\fi
}

\def\wacvPaperID{1550} 
\def\confName{WACV}
\def\confYear{2027}
\title{Reliability-Aware Monocular Depth Supervision \\ for Sparse-View Neural Reconstruction}
\author{
Wei-Teng Chu\textsuperscript{*} \quad Yashasvini Gopalan\textsuperscript{*} \quad Changju Yuan\textsuperscript{*} \\[0.3em]
Stanford University \\
{\tt\small \{waynechu, ygopalan, ycj2003\}@stanford.edu} \\[0.3em]
{\small \textsuperscript{*}Equal contribution}
}
\begin{document}
\maketitle
\begin{abstract}
Sparse-view neural reconstruction in outdoor driving is challenging due to narrow forward-facing trajectories and limited multi-view overlap, and monocular depth priors, though dense, are noisy and not uniformly reliable. We use Depth Anything V2 (DA-V2) as a dense monocular depth prior, align its \Rhi{MRVu}{per-image scale and shift to metric depth using sparse anchors (LiDAR and COLMAP)} and apply depth supervision selectively through photometric masks generated from an RGB-only baseline model, and evaluate on Mip-NeRF-360 and Splatfacto. On KITTISeq02, masked depth supervision gives only marginal gains for Mip-NeRF-360 and does not improve geometry. In contrast, Splatfacto benefits clearly, improving PSNR from 14.903 to 15.932 and reducing RMSE from 0.542 to 0.100. \Rhi{fxQT}{Against global supervision, the proposed mask achieves 0.44--0.70\,dB PSNR gains across KITTI sequences 00/02/05 at tied or better RMSE, while yielding no change on Mip-NeRF-360. This indicates the mask primarily enhances rendering fidelity rather than geometry.} Matched-ratio ablations and \Rhi{fxQT}{two further KITTI fragments} confirm the gains come from selecting reliable low-error regions, rather than from fewer pixels. On the Bicycle scene, depth supervision improves geometry but hurts RGB rendering quality when multi-view coverage is already strong. \Rhi{gpqz}{Using DA-V2 as a representative prior, results suggest that monocular depth priors are valuable for under-constrained sparse-view reconstruction when applied selectively with moderate weighting.}
\end{abstract}

\section{Introduction}
\label{sec:intro}

\begin{figure}[t]
    \centering
    \includegraphics[
        width=\linewidth,
        trim=0 0 25cm 0,
        clip
    ]{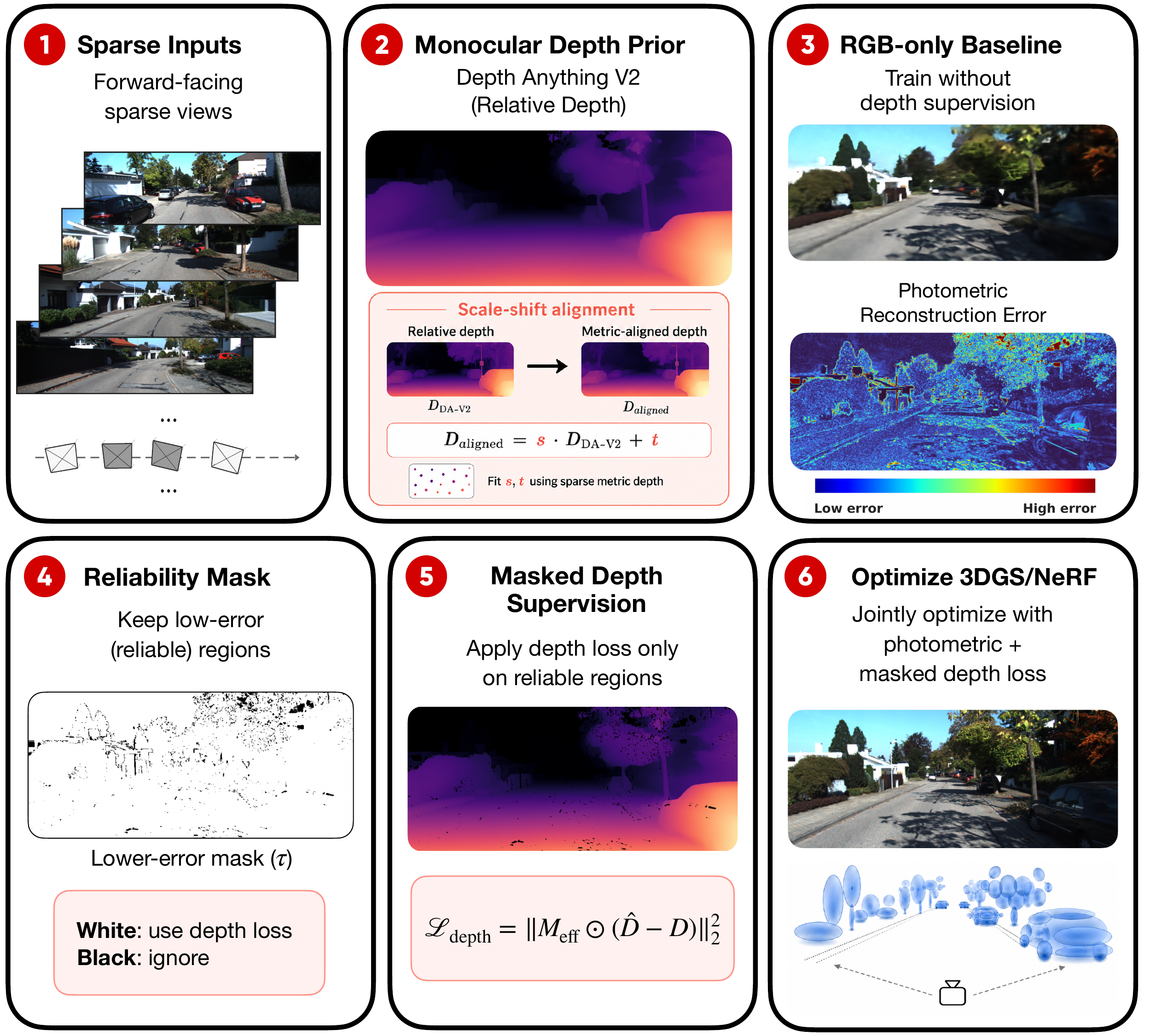}
    \caption{Pipeline overview. DA-V2 depth is scale-shift aligned to metric depth; a photometric reliability mask is computed from an RGB-only baseline; the depth loss is applied only within the low-error mask for joint 3DGS/NeRF optimization.
    }
    \label{fig:full_fig}
\end{figure}

Novel view synthesis aims to reconstruct a scene representation from a set of posed images and render photorealistic images from unseen viewpoints. This problem is central to applications such as autonomous driving simulation, robotics, augmented reality, and digital twins. Recent neural scene representations, including Neural Radiance Fields (NeRF) and 3D Gaussian Splatting (3DGS), have achieved impressive rendering quality by optimizing differentiable scene representations directly from images~\cite{nerf,3dgs}. However, these methods still rely heavily on sufficient multi-view coverage. When the input views are sparse, the reconstruction problem becomes under-constrained, often leading to incorrect geometry and floaters~\cite{ds_nerf,dense_depth_priors,outdoor_nerf_depth}.

This limitation is especially severe in outdoor driving scenes~\cite{kitti,outdoor_nerf_depth}. Unlike object-centric datasets, where cameras move around the target object, autonomous-driving sequences are usually captured by a forward-moving camera along a narrow trajectory. As a result, nearby objects may have limited viewpoint coverage, distant regions provide weak parallax, and large sky regions, shadows, reflective surfaces, and dynamic objects further complicate reconstruction~\cite{outdoor_nerf_depth}. These challenges make sparse-view outdoor reconstruction an important but difficult setting for both NeRF-like and 3DGS-like methods~\cite{mipnerf360,3dgs,outdoor_nerf_depth,in_depth_we_trust}.

Depth supervision is a natural way to reduce this ambiguity. However, accurate ground-truth depth is expensive and often sparse, whereas monocular depth estimators such as Depth Anything V2 (DA-V2) provide dense but imperfect relative depth estimates~\cite{depth_anything_v2}. 

We fit scale and shift to metric depth for DA-V2 predictions and selectively apply depth supervision using photometric reconstruction error from an RGB-only baseline. We evaluate this strategy on two representative scene representations, Mip-NeRF-360 and Splatfacto~\cite{depth_anything_v2,kitti,mipnerf360,nerfstudio,3dgs}. \Rhi{fxQT}{Since the mask modifies an existing depth loss rather than introducing one, we report masked results against \emph{global} (unmasked) supervision, separating the effect of using a depth prior at all from the effect of masking it.}

Our contributions are summarized as follows:
\begin{itemize}
 \Rhi{gpqz} { \item We propose a photometric-error-based masking strategy for monocular depth supervision, validated via matched-ratio ablations that confirm low-error regions provide more reliable depth supervision.}
\item We compare the effect of this strategy on implicit (Mip-NeRF-360) and explicit (Splatfacto) representations.
\item Through matched-ratio mask ablations \Rhi{fxQT}{and aggregation over three KITTI sequences (00/02/05)}, we show that low-photometric-error regions provide more robust supervision for Splatfacto \Rhi{fxQT}{than global unmasked supervision}, while Mip-NeRF-360 remains sensitive to noisy monocular depth.
\Rhi{fxQT}{\item We benchmark the mask against competing depth-supervised methods and a depth-inconsistency mask, outperforming three LiDAR-supervised baselines without any ground-truth depth.}
\Rhi{gpqz} { \item We identify which insights are prior-agnostic versus DA-V2-specific, guiding future depth-prior integrations.}
\item We show depth supervision improves geometry but degrades RGB rendering in object-centric scenes with strong multi-view coverage.
\end{itemize}

\section{Related Work}
\label{sec:related_work}

\paragraph{Neural scene representations.}
NeRF represents a scene as a continuous radiance field and renders images through differentiable volume rendering \cite{nerf}. Mip-NeRF-360 extends it to unbounded outdoor scenes~\cite{mipnerf360}. More recently, 3D Gaussian Splatting represents scenes using explicit anisotropic Gaussians and achieves real-time high-quality rendering \cite{3dgs}. Recent feed-forward Gaussian Splatting methods further use learned multi-view or depth cues for sparse-view reconstruction~\cite{pixelsplat,mvsplat,depthsplat}. We compare both as depth supervision affects implicit and explicit representations differently.

\paragraph{Depth supervision for sparse-view reconstruction.}
Geometric constraints improve sparse-view reconstruction~\cite{ds_nerf,dense_depth_priors,outdoor_nerf_depth}.

\paragraph{Reliability-aware monocular depth priors.}
While depth priors can reduce the geometric ambiguity, monocular depth estimates from recent foundation models~\cite{midas,zoedepth,depth_anything,depth_anything_v2,metric3d,unidepth} are not equally reliable over the entire image because of scale ambiguity, dynamic objects, reflective surfaces, sky regions and occlusion boundaries~\cite{depth_anything_v2,in_depth_we_trust}. Thus, uniform monocular depth supervision may result in wrong geometry in unreliable regions~\cite{in_depth_we_trust}. We focus on when to trust monocular depth supervision~\cite{ds_nerf,dense_depth_priors,outdoor_nerf_depth}. In contrast to prior work that has mostly focused on improving Gaussian Splatting with specific depth losses or inconsistency masks~\cite{depth_regularized_3dgs,dngaussian,sparsegs,dn_splatter,in_depth_we_trust}, we explore a simple photometric-error-based reliability mask and investigate the effect of the same monocular depth prior on both implicit NeRF-style and explicit Gaussian representations. \Rhi{fxQT}{Sec.~\ref{sec:experiments} compares these methods empirically, retrained on our sparse forward-facing setting, and benchmarks our mask against a static approximation of the inconsistency mask of~\cite{in_depth_we_trust}.}

\section{Data}
\label{sec:data}

We evaluate our approach on two benchmark datasets, selected sequences from the KITTI  \cite{kitti} odometry benchmark and the outdoor object-centric \textit{Bicycle} scene from Mip-NeRF-360 \cite{mipnerf360}  dataset.

\textbf{KITTI.} KITTI comprises large-scale real-world outdoor driving scenes, making it well suited for evaluating novel view synthesis in autonomous driving contexts. Unlike object-centric scenes typical of standard NeRF and 3DGS benchmarks, vehicles in autonomous driving scenarios exhibit predominantly forward or turning motion along a fixed trajectory. To mitigate the influence of lighting variation and dynamic objects, we follow the broader evaluation protocol of prior outdoor NeRF work~\cite{outdoor_nerf_depth}, which draws from selected KITTI odometry sequences 00, 02, and 05. For each sequence, every 10th frame is held for the test set and the remaining frames are used for training. To assess robustness under sparse viewpoints, we simulate low-frequency imaging at 2.5 Hz by subsampling 50\% of the training frames (retaining every second frame). We focus on KITTI sequence 02 (fragment 034) as the main evaluation setting, with \Rhi{fxQT}{sequence 05 (fragment 018) and sequence 00 (fragment 027) as generalization checks}. \Rhi{fxQT}{Fragment 034 spans 88 frames (every 2nd of a 175-frame raw sequence) with 8 test frames at the every-10th sparse index; fragments 018 and 027 follow the identical protocol.} Rather than relying on Structure-from-Motion (SfM) poses, which may introduce scale inconsistencies with ground-truth depth, we use the calibrated poses provided directly by the KITTI odometry benchmark.

\textbf{Mip-NeRF-360.} We further evaluate on the outdoor \textit{Bicycle} scene from the Mip-NeRF-360 dataset, using 194 sparse views. Every 10th frame is held back for validation and testing. To accommodate GPU memory constraints, images are downscaled by 4× for 3DGS training, while depth priors are computed at full resolution and resized at load time.

\begin{figure}
\centering
\includegraphics[width=1\linewidth]{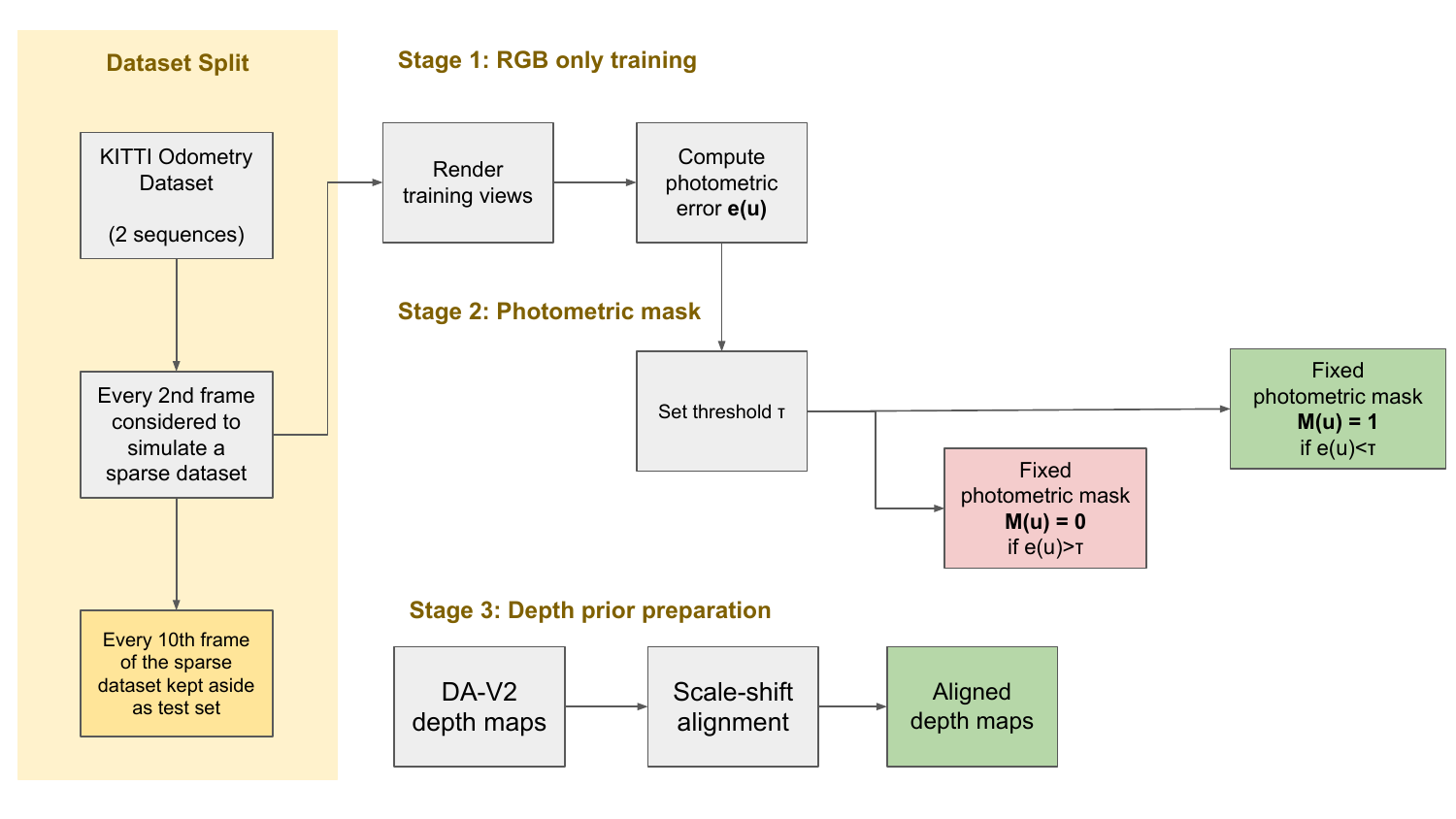}
\caption{Training views are subsampled from the KITTI Odometry dataset to simulate sparse-view reconstruction. Fixed photometric masks are generated from baseline renderings using a threshold on the photometric error, while DA-V2 predictions are scale-shift aligned to produce metric depth priors for supervision.}
\label{fig:placeholder}
\end{figure}

\section{Methods}
\label{sec:methods}

\subsection{Overview}

Photometric supervision alone under-constrains sparse-view reconstruction, and monocular depth estimates are unreliable in regions such as dynamic objects, reflective surfaces, and sky. We therefore apply DA-V2 depth supervision selectively rather than uniformly.

Our approach has three steps: (1) align each DA-V2 map to metric depth via sparse anchors (LiDAR in KITTI, COLMAP in \textit{Bicycle}); (2) train an RGB-only baseline and estimate per-pixel photometric error; (3) retrain with a masked depth loss that supervises only low-error pixels. Both Splatfacto and Mip-NeRF-360 share the same depth prior, mask, and masked-depth objective.

\subsection{Scale-Aligned Monocular Depth Prior}

We use DA-V2 to predict a dense monocular depth map for each training image. Since DA-V2 predicts relative depth rather than metric depth, its raw output cannot be directly used as supervision. We therefore align each predicted depth map to a metric reference before training. For KITTI, the reference comes from valid LiDAR depth pixels. For the Mip-NeRF-360 Bicycle scene, we use sparse COLMAP points as anchor depth.

Given a monocular prediction ($d_m(u)$) and reference depth ($d_r(u)$) at valid anchor pixels ($u \in \Omega$), we solve a per-image scale-shift alignment:
\begin{equation}
s^*, t^* = \arg\min_{s,t} \sum_{u \in \Omega} \left(s d_m(u) + t - d_r(u)\right)^2 .
\end{equation}

The aligned depth prior is then
\begin{equation}
d(u) = s^* d_m(u) + t^* .
\end{equation}

For KITTI, we clip the aligned depth to the valid evaluation range of 80 meters. Pixels without valid reference depth are not used to estimate the scale and shift, but the resulting aligned dense map is used as a depth prior during training.

\Rhi {MRVu} {While DA-V2 provides dense relative structure, absolute metric scale recovery relies entirely on sparse anchors. Consequently, downstream task performance remains bounded by anchor density and accuracy.}

\subsection{Photometric Reliability Mask}

We investigate the effects of a masked depth prior, in which we first estimate where the RGB reconstruction is reliable. Given a rendered image $\hat{I}$ from a depth-free baseline model and a ground-truth image $I$, we compute the photometric error per-pixel:
\begin{equation}
e(u) = \frac{1}{3} \sum_{c \in \{R,G,B\}}
\left| \hat{I}_c(u) - I_c(u) \right|,
\end{equation}
where $u$ denotes a pixel location and $e(u)\in[0,1]$ for normalized RGB images.

Pixels are then classified using a threshold $\tau$. We focus primarily on low-error masking ($e(u) < \tau$), where depth supervision is applied only to pixels that are accurately reconstructed by the RGB model.

We define the binary photometric reliability mask as

\begin{equation}
M_{\tau}(u) =
\begin{cases}
1, & e(u) < \tau, \\
0, & \text{otherwise}.
\end{cases}
\end{equation}

The motivation for this mask is that large photometric error often indicates regions where supervision is less reliable, such as dynamic objects, reflective surfaces, occlusion boundaries, or sky regions. Applying a monocular depth loss to these pixels may force the model toward an inaccurate geometry. In contrast, low-error pixels are more likely to correspond to regions that are already consistent with the RGB observations, so the aligned monocular depth prior can act as a stabilizing geometric regularizer. \Rhi{shared}{Photometric and depth error are measured against different references, so low photometric error does not by construction imply low depth error. We therefore validate the mask against ground-truth LiDAR before using it (Sec.~\ref{sec:experiments}).}

During depth-supervised training, the photometric reliability mask is fixed after being calculated once from the RGB-only baseline.

\subsection{Masked Depth Supervision Objective}

Let $M_{\tau}(u)$ denote the fixed photometric reliability mask and $D(u)$ denote the depth-validity mask. The effective supervision mask is
\begin{equation}
M_{\mathrm{eff}}(u) = M_{\tau}(u) \land D(u).
\end{equation}

The depth loss is computed only on pixels selected by the effective supervision mask:
\begin{equation}
\mathcal{L}_{\mathrm{depth}} =
\frac{1}{N}
\sum_{u} M_{\mathrm{eff}}(u)
\left( \hat{d}(u) - d(u) \right)^2,
\end{equation}
where $\hat{d}(u)$ is the rendered depth and $d(u)$ is the aligned depth prior.

The overall training objective is
\begin{equation}
\mathcal{L} = \mathcal{L}_{\text{rgb}} + \lambda_{\text{depth}}\, \mathcal{L}_{\text{depth}},
\end{equation}
where $\lambda_{\mathrm{depth}}$ controls the relative weight of depth supervision. The RGB reconstruction loss is evaluated over the full image, while the depth loss is applied only within the effective supervision mask.
\Rhi{fxQT}{Setting $\tau{=}1.0$ gives $M_{\tau}\equiv 1$, i.e.\ global (unmasked) supervision. This condition, not the depth-free $\lambda_{\mathrm{depth}}{=}0$ baseline, isolates the effect of the mask.}

\subsection{Training Pipeline}

Our approach consists of three stages. First, we train a baseline reconstruction model without depth supervision ($\lambda_{\mathrm{depth}} = 0$) for 50,000 iterations. The resulting model is used to render all training views and estimate per-pixel photometric error. We then set a threshold for the error maps to generate binary photometric masks, which remain fixed for the remainder of training. Finally, the model is retrained \Rhi{gpqz} {for another 50,000 iterations from a fresh initialization} with depth supervision enabled while applying the photometric masks to gate the depth loss. The threshold determines the fraction of pixels that contribute to depth supervision, while $\lambda_{\mathrm{depth}}$ controls the relative weight of the depth loss. \Rhi{gpqz} {Both parameters were selected via an empirical grid search over the validation data.}

\subsection{Backbone-Specific Implementation}

\paragraph{Mip-NeRF-360.}
For Mip-NeRF-360, depth supervision is incorporated through a mean squared error (MSE) loss between the rendered depth and the aligned depth prior. Pixels excluded by the fixed photometric mask are marked as invalid supervision and do not contribute to the depth loss. The RGB rendering objective remains unchanged.

\paragraph{Splatfacto.}
We extend Nerfstudio's Splatfacto \cite{splatfacto} implementation to load aligned DA-V2 depth maps through the dataset configuration. During training, depth loss is computed only on pixels that satisfy both depth-validity and photometric-mask constraints. The RGB rendering objective remains the same.

\section{Experiments}
\label{sec:experiments}

\subsection{Experimental Setup}
We evaluate on KITTISeq02 (every-2 sparse-view), with \Rhi{fxQT}{KITTISeq05 and KITTISeq00 as generalization checks.} We compare Mip-NeRF-360 (implicit) and Splatfacto (explicit)~\cite{mipnerf360,3dgs,nerfstudio}; DA-V2 depth maps are scale-shift aligned to KITTI and clipped at 80\,m~\cite{depth_anything_v2}. We sweep $\tau \in \{0.14,0.16,0.18,0.20,0.22,1.0\}$ and $\lambda \in \{0,0.05,0.10,0.15\}$; \Rhi{fxQT}{$\tau{=}1.0$ is global (unmasked) supervision, the primary baseline for judging the mask.} Matched-ratio ablations at $\tau{=}0.18$, $\lambda{=}0.10$ compare low-, high-, and random-error masks at equal pixel count. \Rhi{fxQT}{Best settings are retrained over 3 seeds (mean$\pm$std); all other numbers are single runs.} Rendering: PSNR, SSIM, LPIPS~\cite{ssim,lpips}; geometry: AbsRel and RMSE. We also test Splatfacto on the \textit{Bicycle} scene.

\subsection{Depth Prior Quality}
We evaluate the aligned DA-V2 maps on valid KITTI depth pixels before using DA-V2 as supervision~\cite{kitti,depth_anything_v2}. The average absolute error of alignment is 4.22 m. \Rhi{MRVu}{To test sensitivity to anchor density, we subsample the LiDAR points before optimization (Table~\ref{tab:anchor_density}). The alignment MAE remains stable from $\sim$95k anchors down to 500 per frame, only degrading under extreme sparsity (100 anchors: 4.29\,m; 20 anchors: 4.51\,m). This confirms that the 2-DOF optimization saturates long before typical LiDAR density is reached. The 4.22\,m error floor reflects DA-V2's intrinsic local structural limitations rather than calibration artifacts.}

\Rhi{shared}{Table~\ref{tab:mask_validity} tests the mask directly. Depth RMSE is lower inside the mask at every threshold (34--39\% reduction; 6.33 vs.\ 10.01\,m at $\tau{=}0.18$). Inside-mask RMSE increases monotonically across all six $\tau$ values ($r{=}1.0$); outside-mask RMSE likewise increases across the five thresholds where it is defined ($\tau{=}0.14$--$0.22$, $r{=}1.0$). AbsRel drops only 10.0\% ($r{=}0.029$, $p{=}0.96$), consistent with a depth-range confound. The mask therefore selects pixels where the prior is more accurate in absolute terms, which is what the depth loss uses, rather than uniformly better relative depth. Tables~\ref{tab:mask-validation} and~\ref{tab:mask-significance} give the per-threshold breakdown and significance tests.}

\begin{table}[h]
\centering\small
\begin{tabular}{ccc}
\toprule
$\tau$ & \shortstack{Inside \\ RMSE$\downarrow$ (m)} & \shortstack{Outside \\ RMSE$\downarrow$ (m)} \\
\midrule
0.14 & 6.30 & 9.58 \\
0.16 & 6.32 & 9.82 \\
\textbf{0.18} & \textbf{6.33} & \textbf{10.01} \\
0.20 & 6.36 & 10.24 \\
0.22 & 6.38 & 10.45 \\
1.00 & 6.43 & — \\
\bottomrule
\end{tabular}
\caption{\Rhi{shared}{DA-V2 depth RMSE inside vs.\ outside the mask (KITTISeq02). 34--39\% lower inside; both columns monotone ($r{=}1.0$). \textbf{Bold}: $\tau{=}0.18$.}}
\label{tab:mask_validity}
\end{table}

\begin{table}[h]
\centering\small
\begin{tabular}{cc}
\toprule
Anchors/frame & MAE (m) \\
\midrule
$\sim$95{,}000 & 4.22 \\
5{,}000  & 4.22 \\
500      & 4.22 \\
100      & 4.29 \\
50       & 4.29 \\
20       & 4.51 \\
\bottomrule
\end{tabular}
\caption{\Rhi{MRVu}{Alignment MAE vs.\ anchor count (KITTISeq02, 88 training frames). Stable to $\sim$500/frame. Degrades only under extreme sparsity.}}
\label{tab:anchor_density}
\end{table}

\subsection{Mip-NeRF-360 Results}

\noindent\textbf{Takeaway.} Masked monocular depth supervision provides only weak and unstable regularization for Mip-NeRF-360: it slightly improves PSNR in some KITTISeq02 settings, but does not improve SSIM, LPIPS, or metric geometry. On KITTISeq05, both global and masked depth supervision degrade rendering and geometry compared with RGB-only training. \Rhi{fxQT}{Against global supervision the mask gains nothing: global attains the best KITTISeq02 PSNR (20.607), and on KITTISeq05 masking only reduces global's degradation.}

Table~\ref{tab:mipnerf_results_compact} summarizes Mip-NeRF-360 results on KITTISeq02~\cite{mipnerf360,kitti}.

\begin{figure*}[!htbp]
\centering
\setlength{\tabcolsep}{1.5pt}
\renewcommand{\arraystretch}{0.3}
\footnotesize

\begin{tabular}{@{}c@{\hspace{4pt}}cccccc@{}}
 & $\tau{=}0.14$ & $\tau{=}0.16$ & $\tau{=}0.18$ & $\tau{=}0.20$ & $\tau{=}0.22$ & global \\

\rotatebox{90}{$\lambda{=}0.05$} &
\includegraphics[width=.155\linewidth]{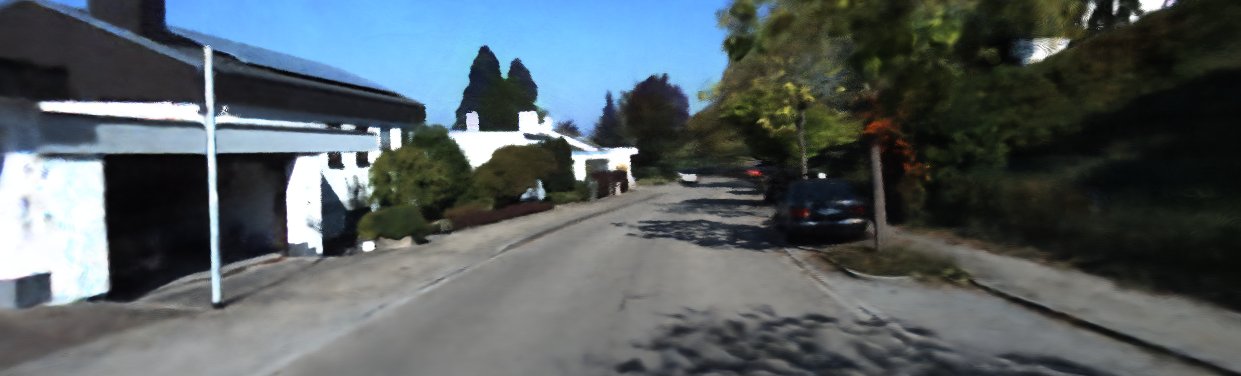} &
\includegraphics[width=.155\linewidth]{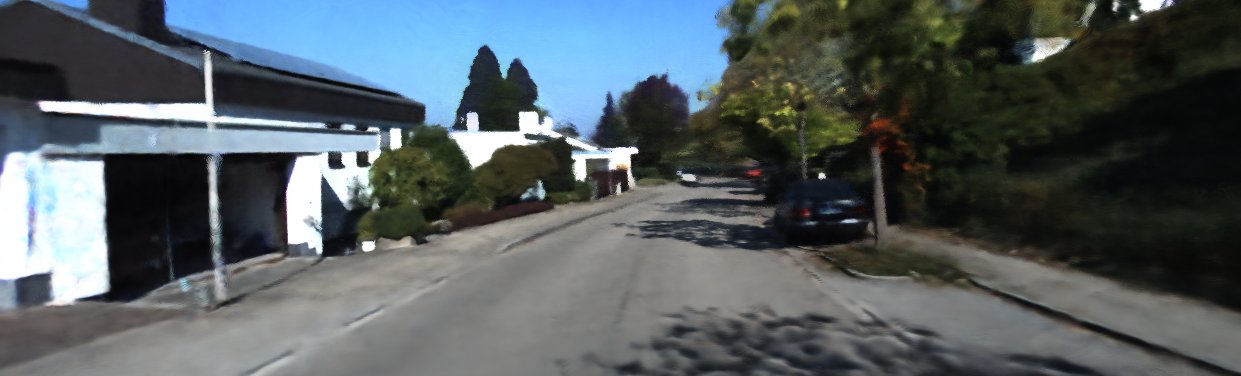} &
\includegraphics[width=.155\linewidth]{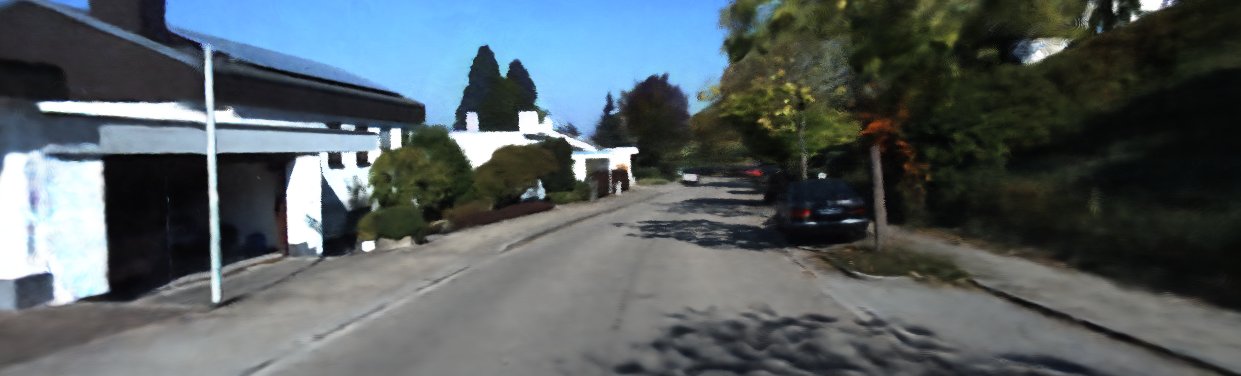} &
\includegraphics[width=.155\linewidth]{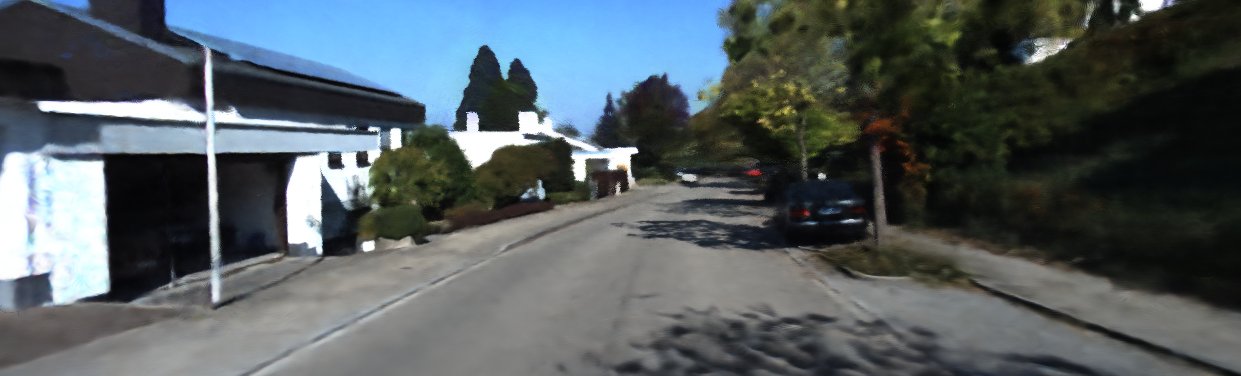} &
\includegraphics[width=.155\linewidth]{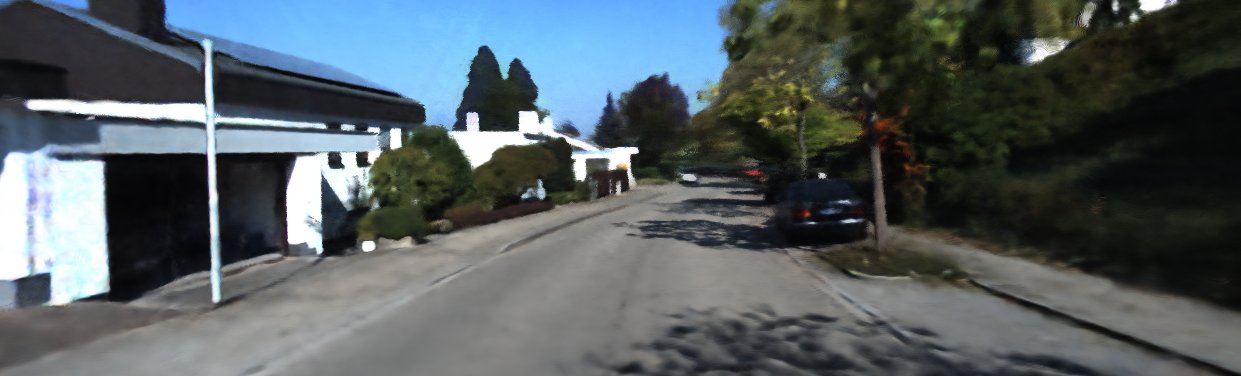} &
\includegraphics[width=.155\linewidth]{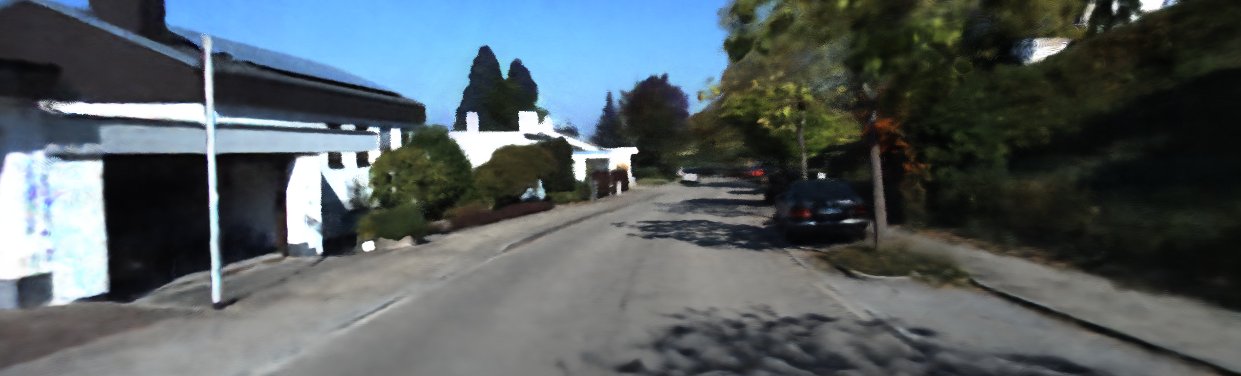} \\

\rotatebox{90}{$\lambda{=}0.10$} &
\includegraphics[width=.155\linewidth]{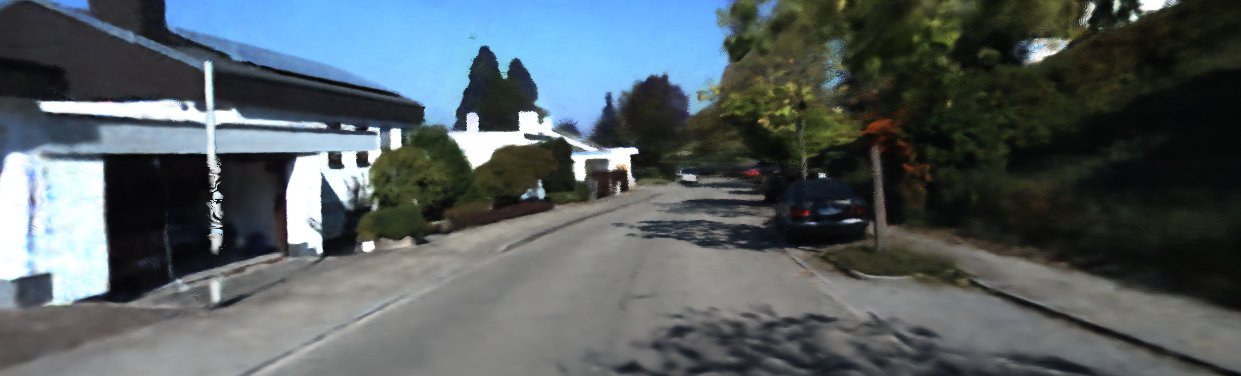} &
\includegraphics[width=.155\linewidth]{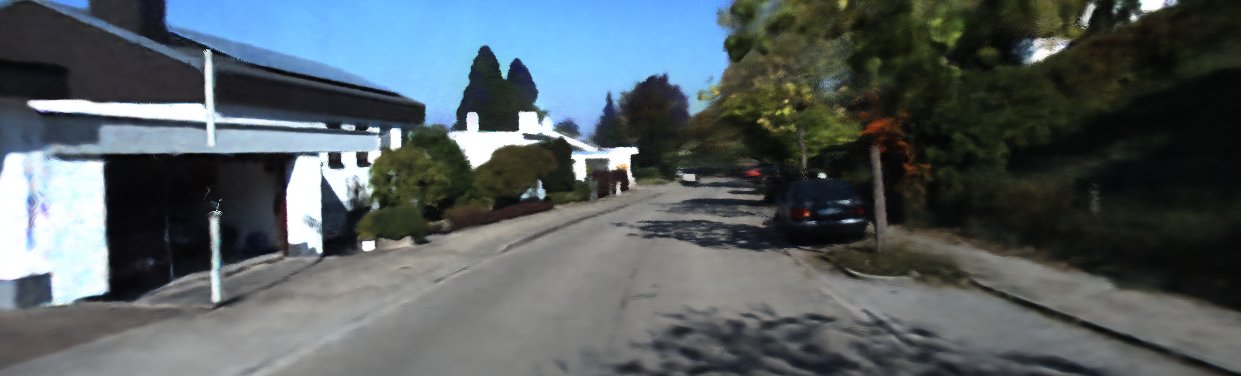} &
\includegraphics[width=.155\linewidth]{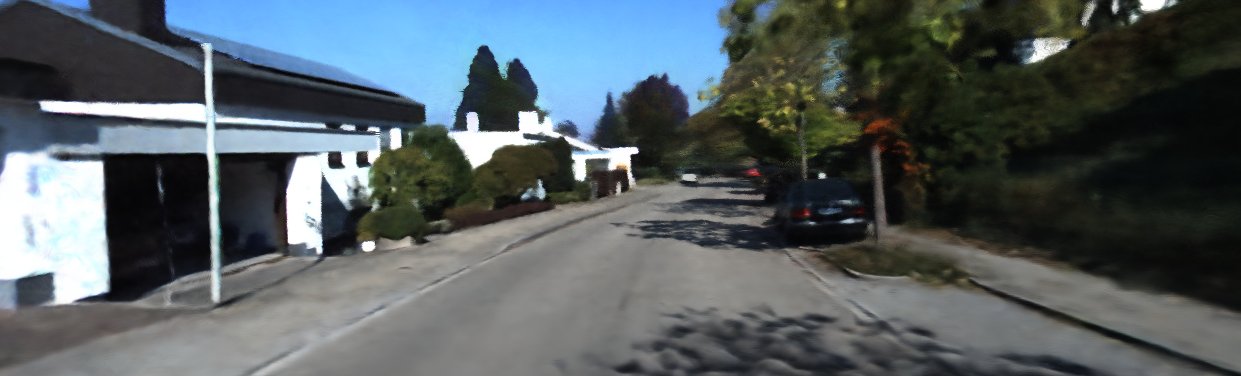} &
\includegraphics[width=.155\linewidth]{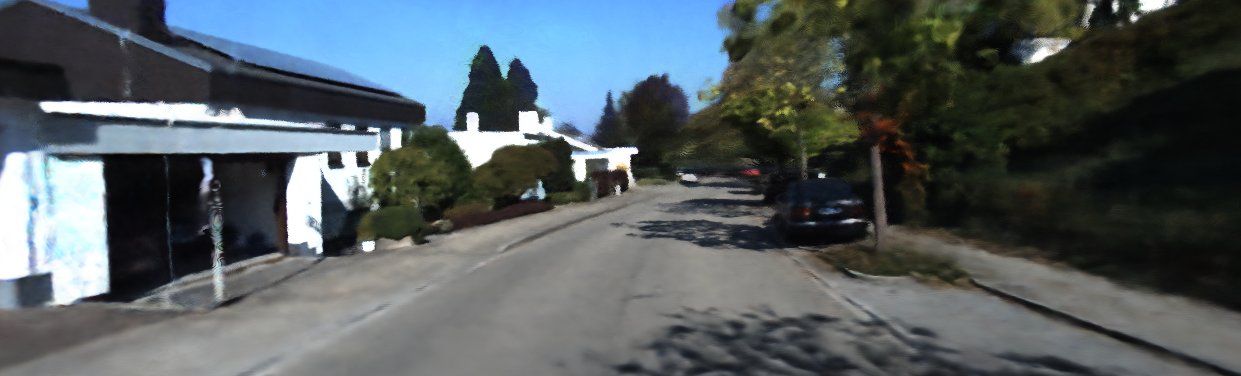} &
\includegraphics[width=.155\linewidth]{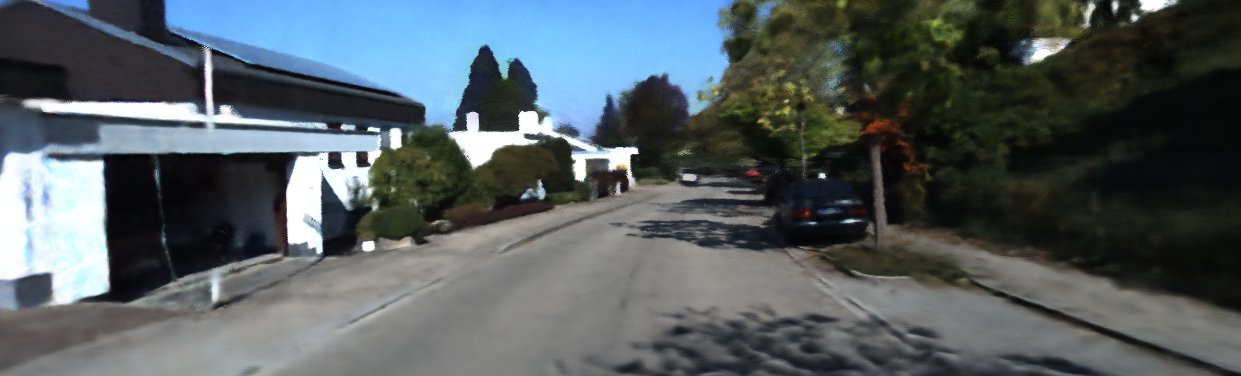} &
\includegraphics[width=.155\linewidth]{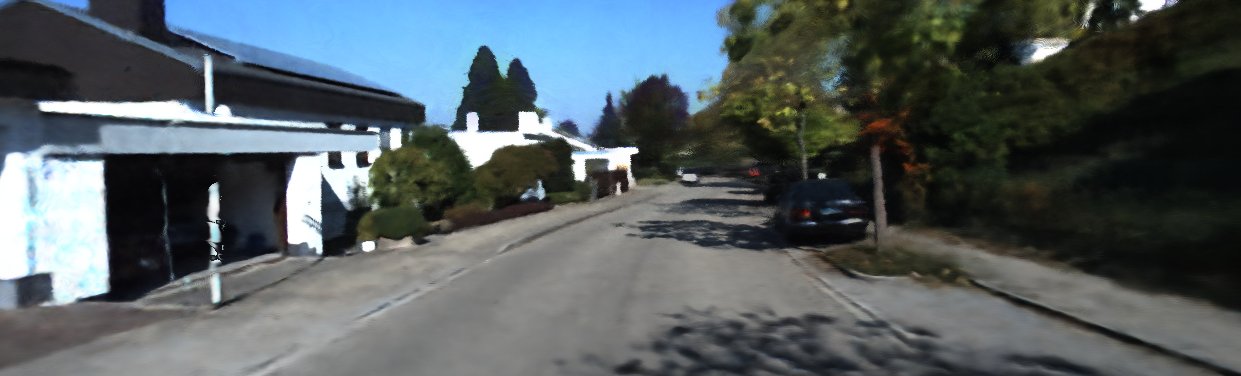} \\

\rotatebox{90}{$\lambda{=}0.15$} &
\includegraphics[width=.155\linewidth]{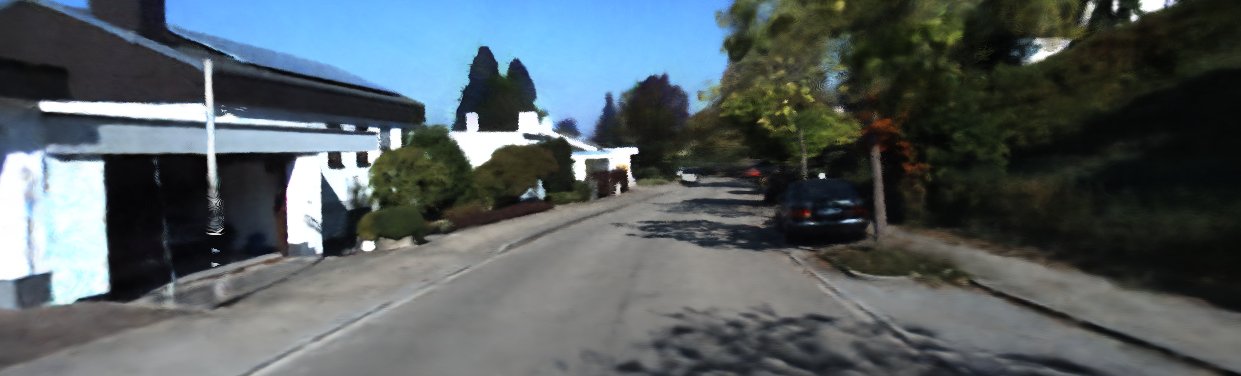} &
\includegraphics[width=.155\linewidth]{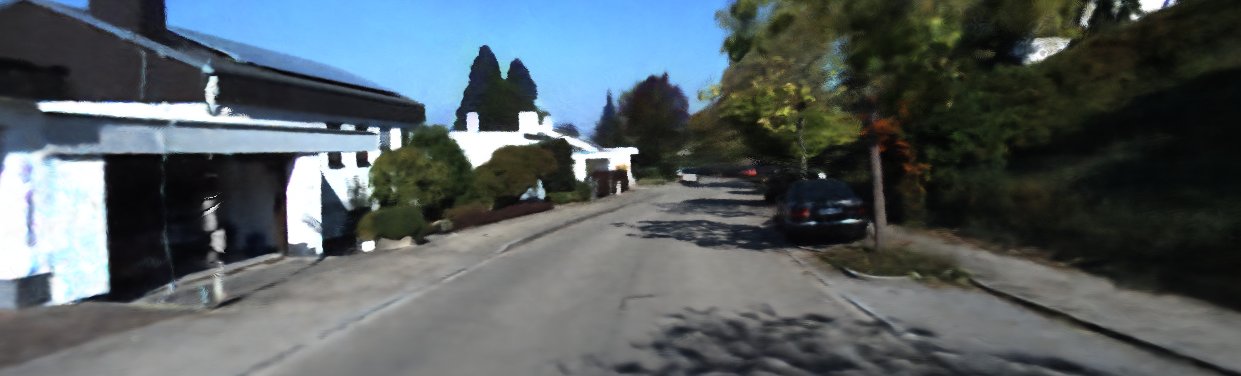} &
\includegraphics[width=.155\linewidth]{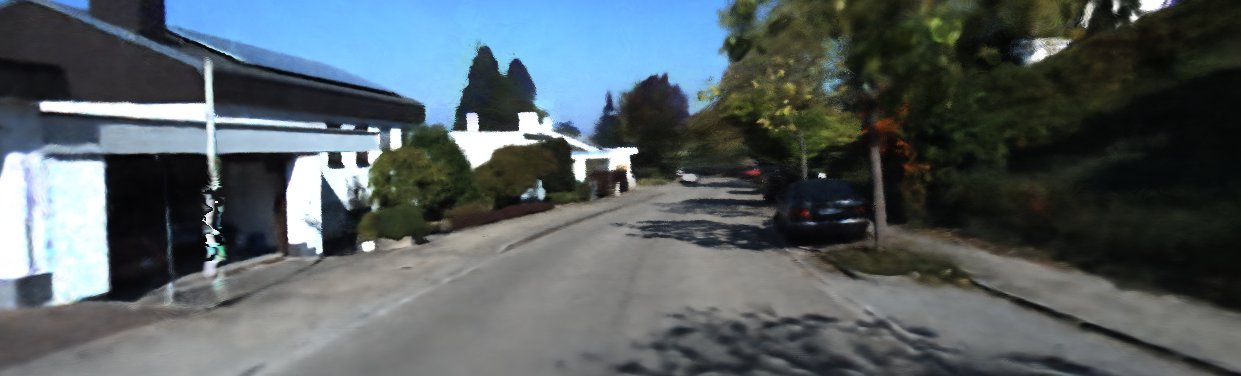} &
\includegraphics[width=.155\linewidth]{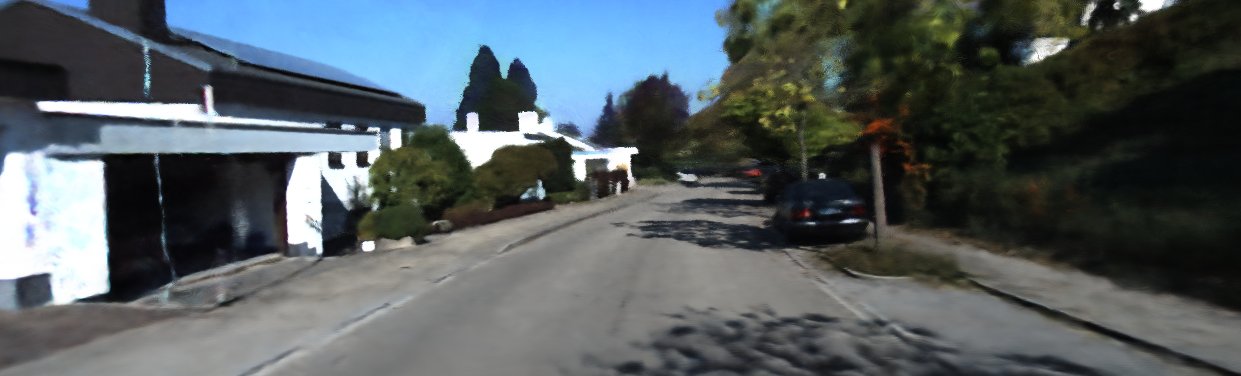} &
\includegraphics[width=.155\linewidth]{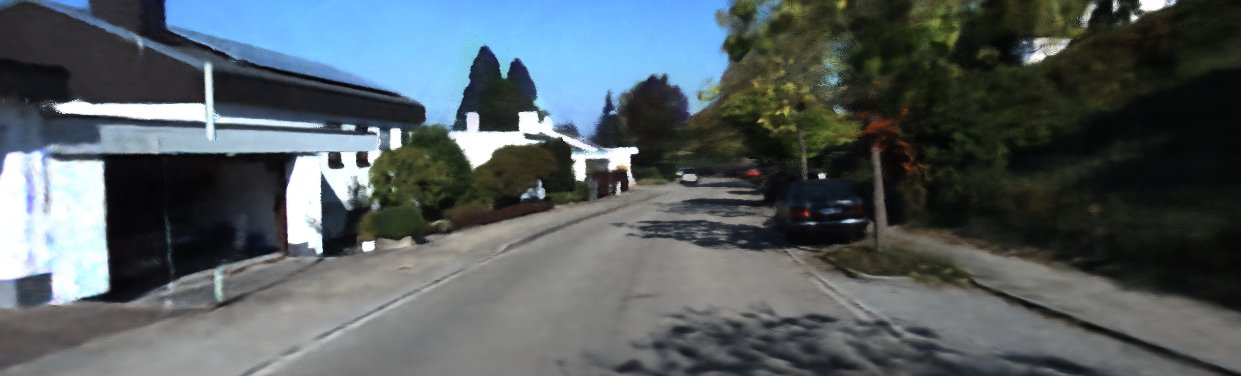} &
\includegraphics[width=.155\linewidth]{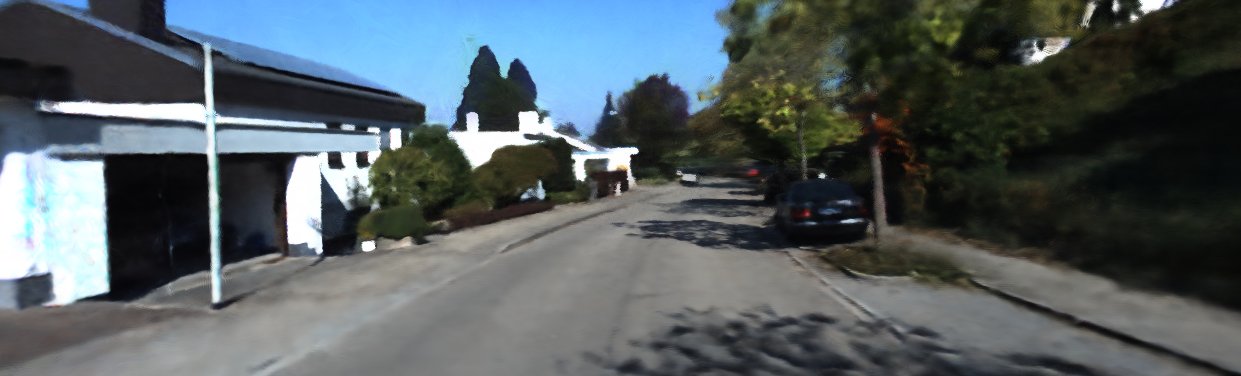} \\
\end{tabular}

\vspace{6pt}

\begin{tabular}{@{}cc@{\hspace{20pt}}c@{}}
\includegraphics[width=.22\linewidth]{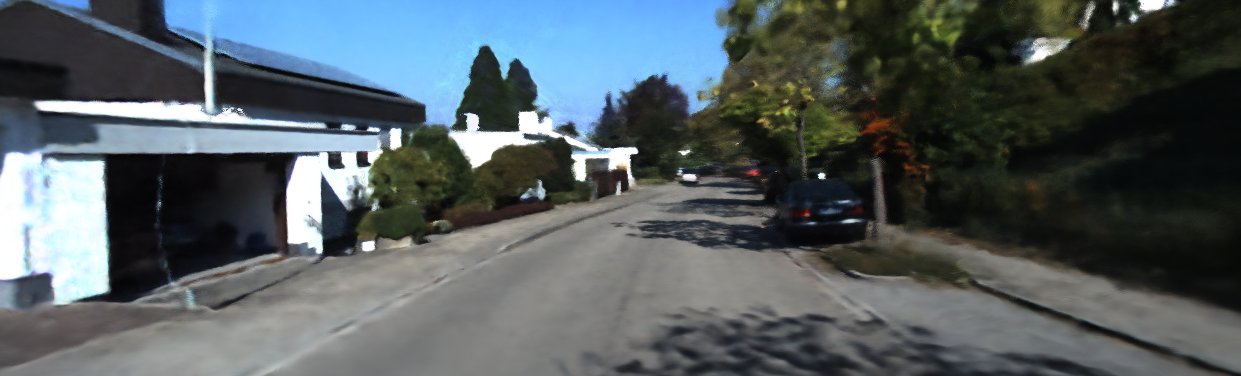} & &
\includegraphics[width=.22\linewidth]{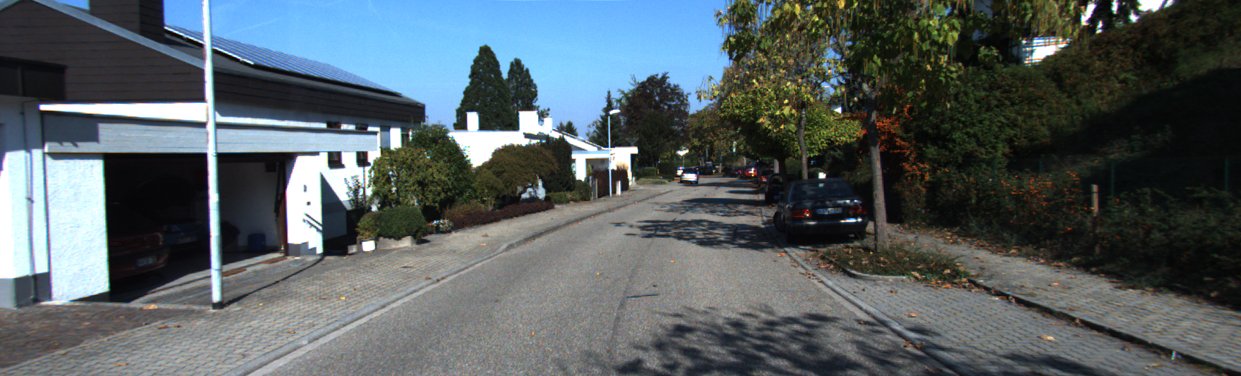} \\
$\lambda{=}0$ (no depth prior) & & Ground Truth \\
\end{tabular}

\caption{Ablation on depth loss weight $\lambda$ and photometric reliability threshold $\tau$ shows effect on Mip-NeRF-360 reconstruction of finer details such as the street pole. }
\label{fig:ablation_mipnerf}
\end{figure*}

\begin{figure*}[!htbp]
\centering
\setlength{\tabcolsep}{1.5pt}
\renewcommand{\arraystretch}{0.3}
\footnotesize

\begin{tabular}{@{}c@{\hspace{4pt}}cccccc@{}}
 & $\tau{=}0.14$ & $\tau{=}0.16$ & $\tau{=}0.18$ & $\tau{=}0.20$ & $\tau{=}0.22$ & global \\

\rotatebox{90}{$\lambda{=}0.05$} &
\includegraphics[width=.155\linewidth]{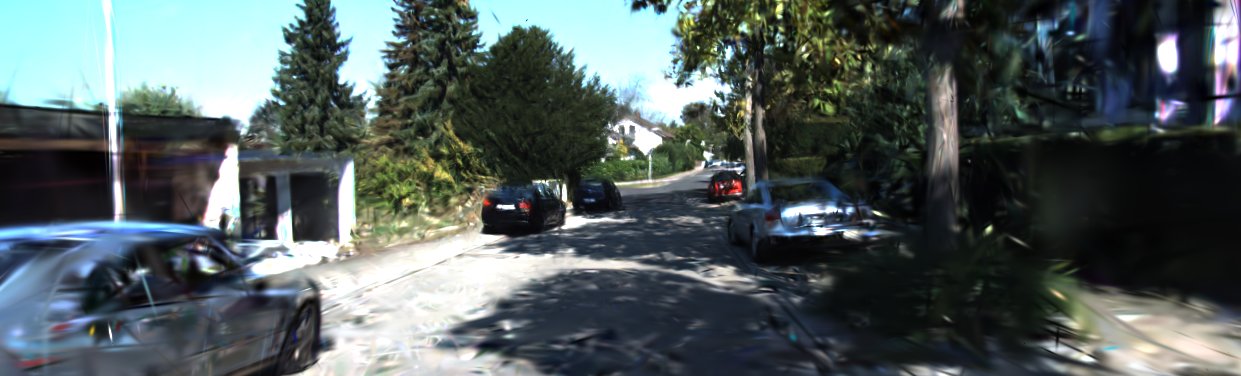} &
\includegraphics[width=.155\linewidth]{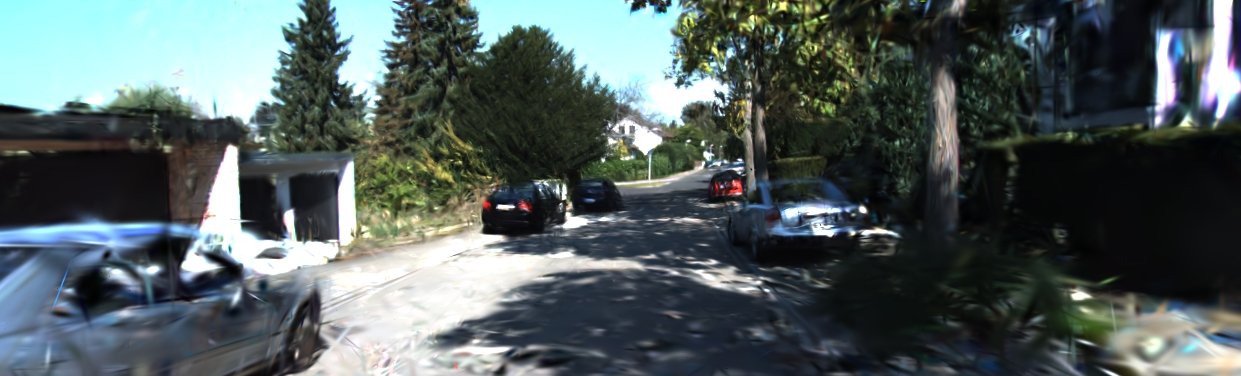} &
\includegraphics[width=.155\linewidth]{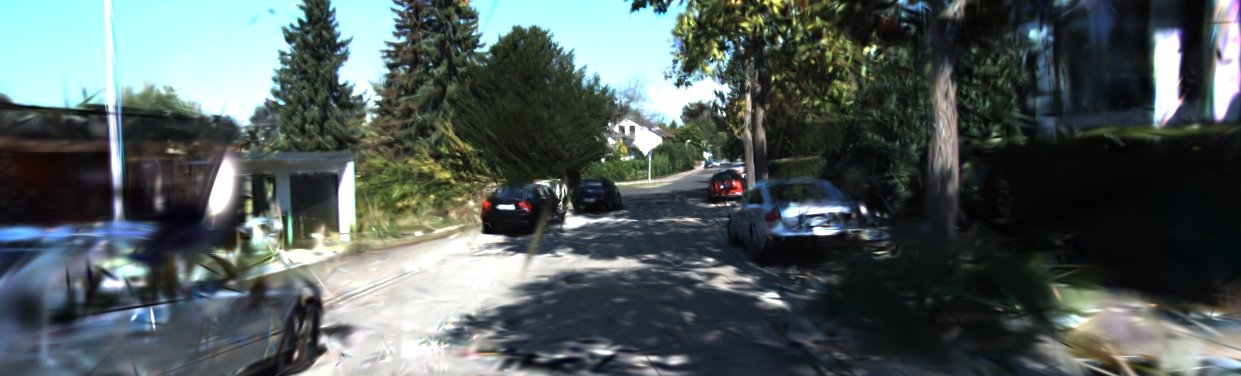} &
\includegraphics[width=.155\linewidth]{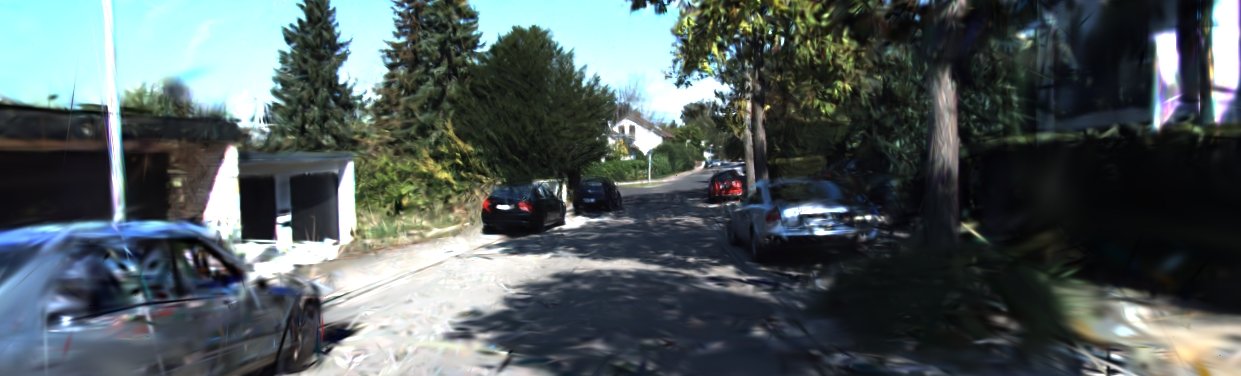} &
\includegraphics[width=.155\linewidth]{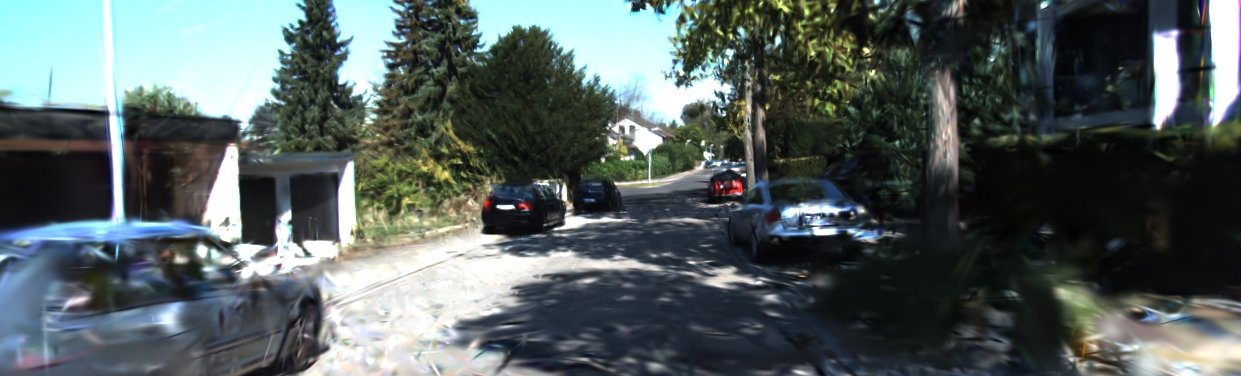} &
\includegraphics[width=.155\linewidth]{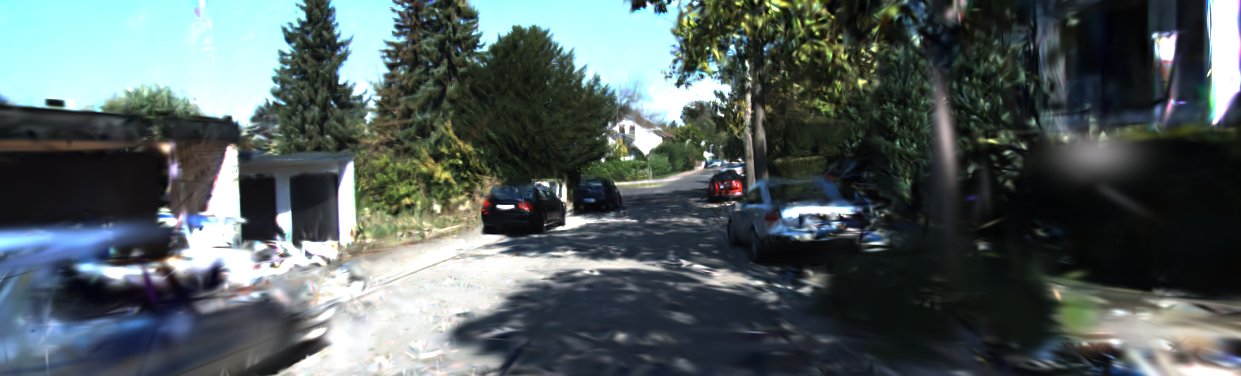} \\

\rotatebox{90}{$\lambda{=}0.10$} &
\includegraphics[width=.155\linewidth]{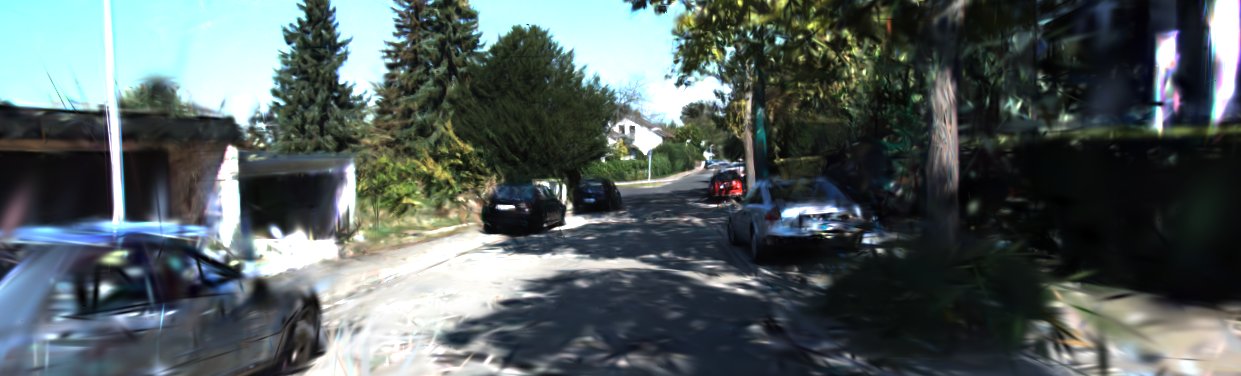} &
\includegraphics[width=.155\linewidth]{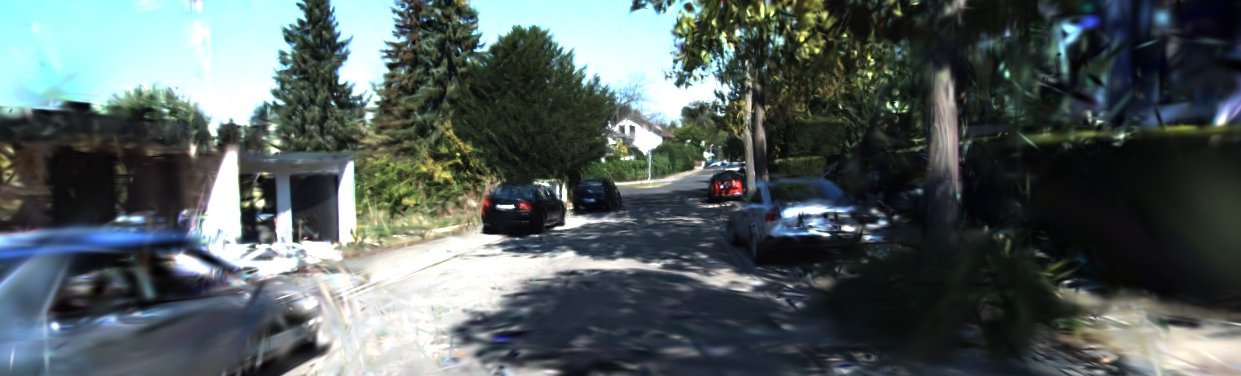} &
\includegraphics[width=.155\linewidth]{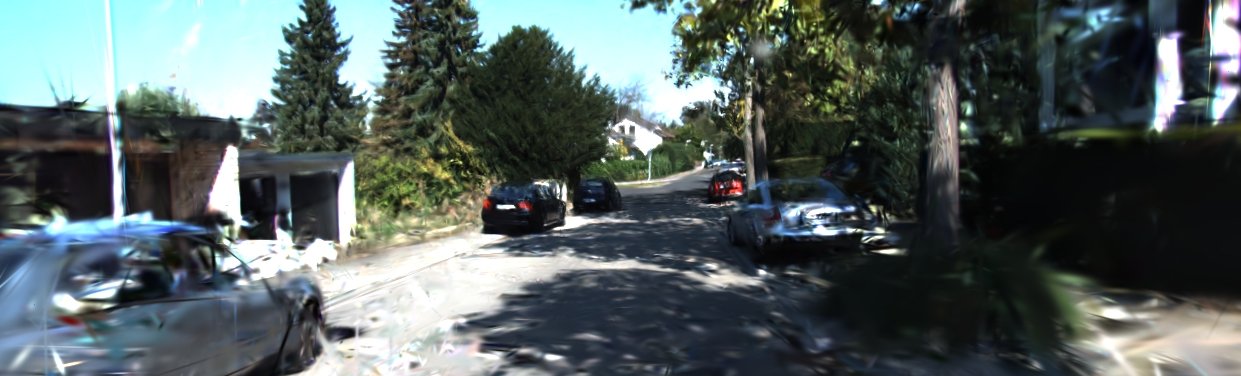} &
\includegraphics[width=.155\linewidth]{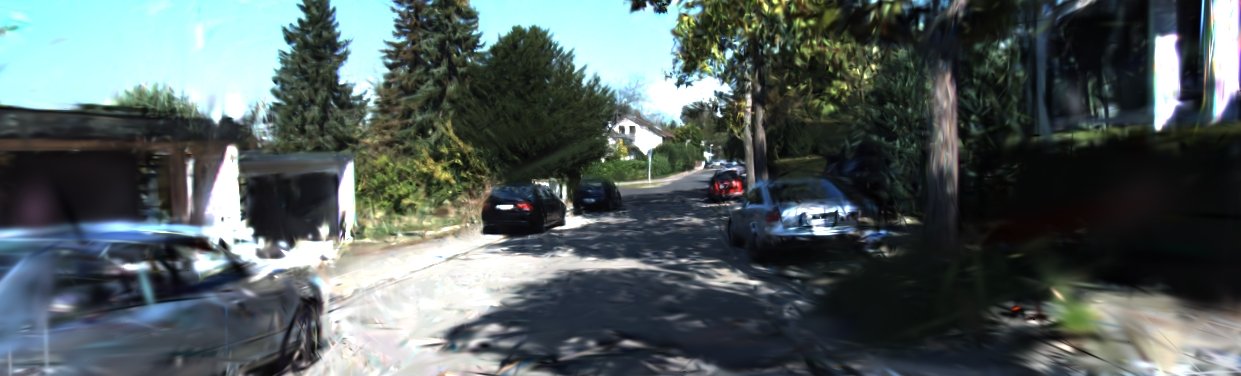} &
\includegraphics[width=.155\linewidth]{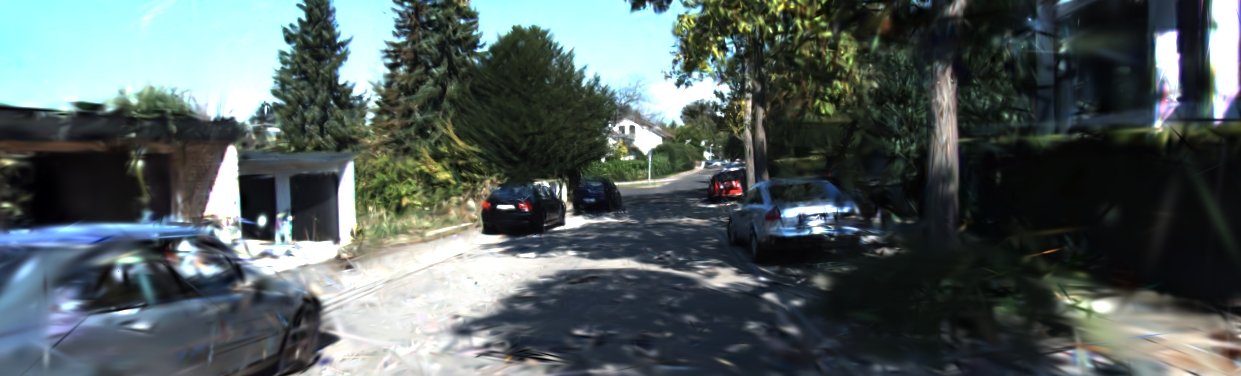} &
\includegraphics[width=.155\linewidth]{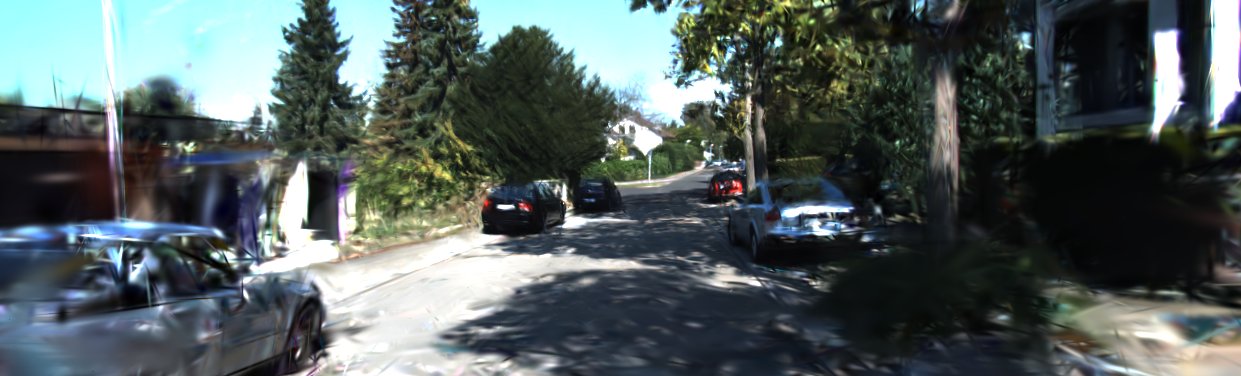} \\

\rotatebox{90}{$\lambda{=}0.15$} &
\includegraphics[width=.155\linewidth]{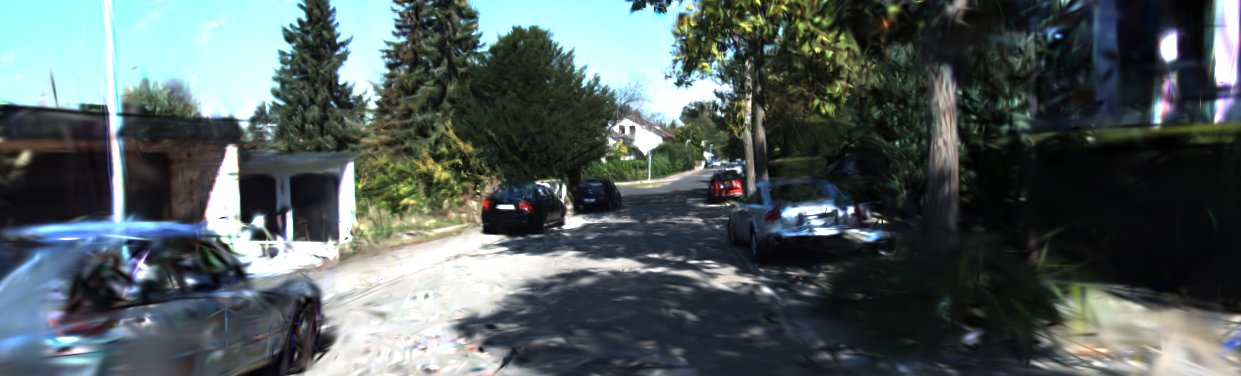} &
\includegraphics[width=.155\linewidth]{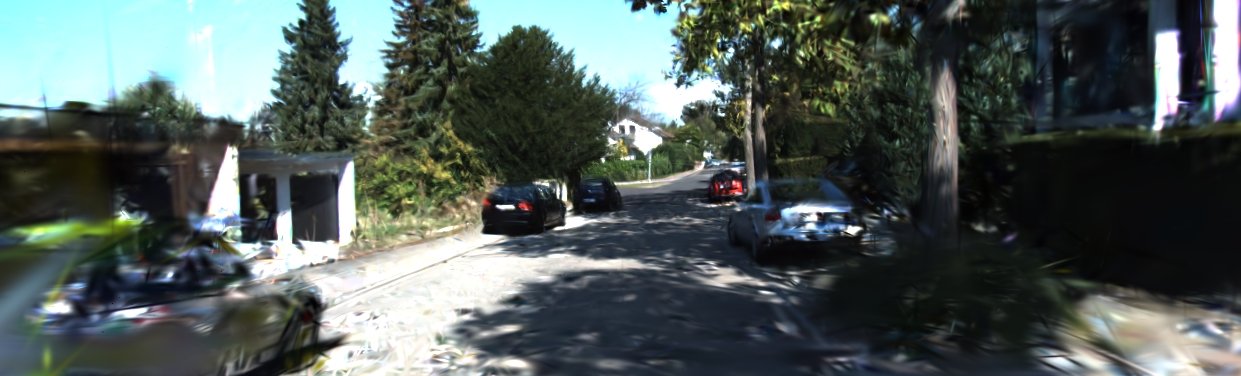} &
\includegraphics[width=.155\linewidth]{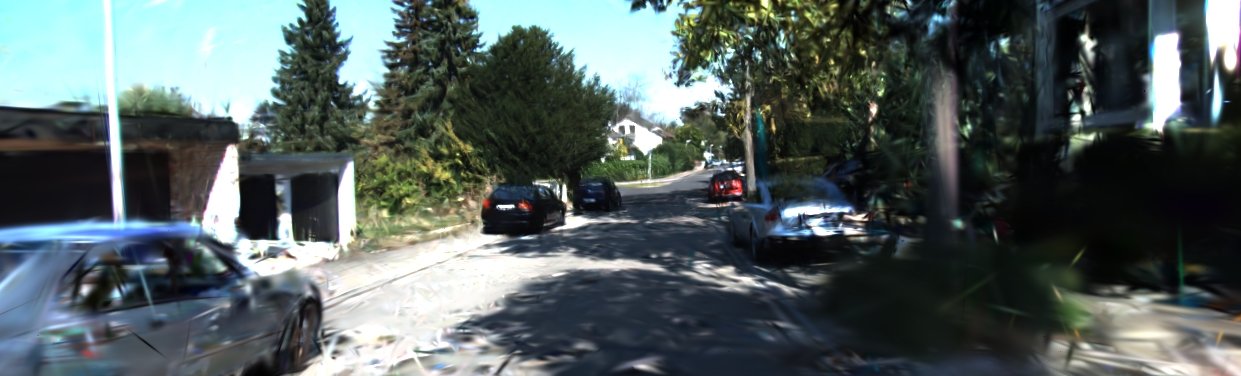} &
\includegraphics[width=.155\linewidth]{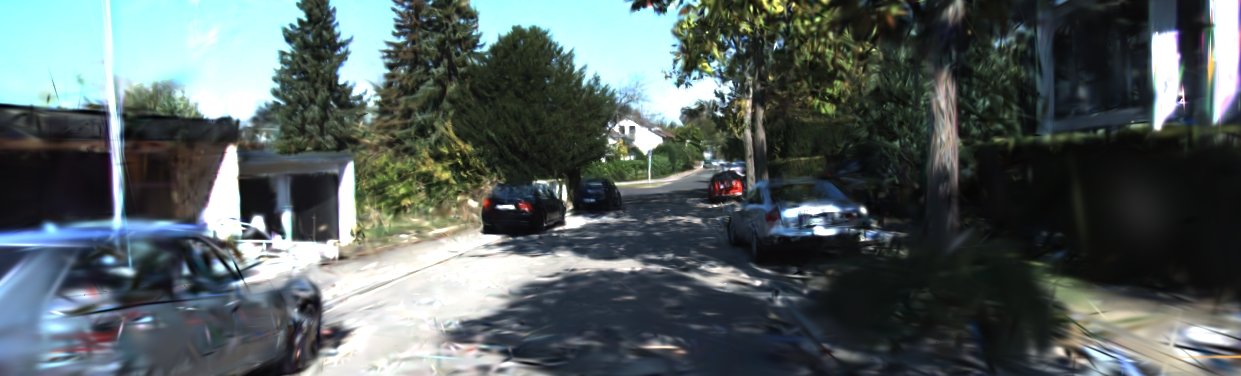} &
\includegraphics[width=.155\linewidth]{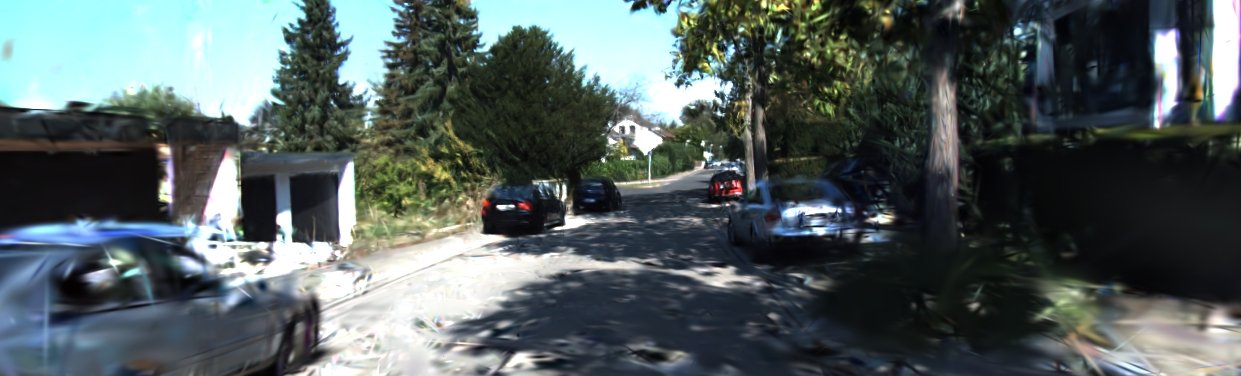} &
\includegraphics[width=.155\linewidth]{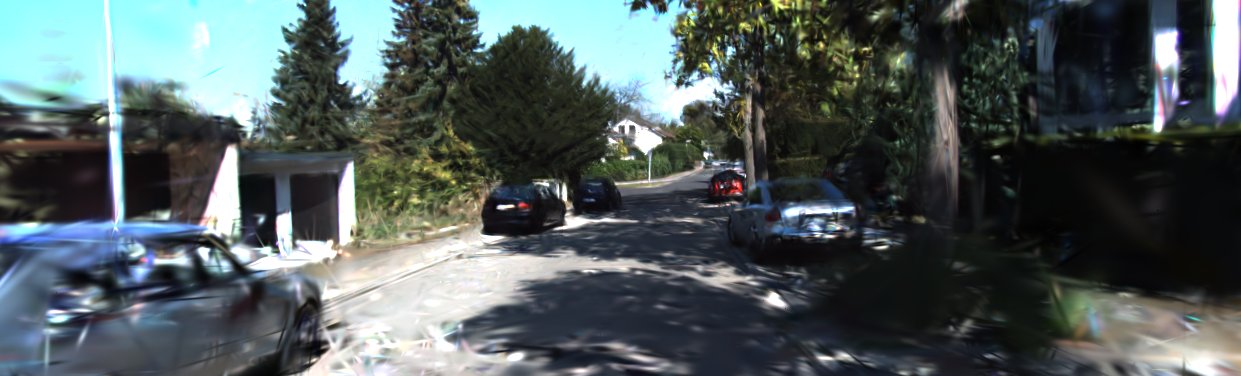} \\
\end{tabular}

\vspace{6pt}

\begin{tabular}{@{}cc@{\hspace{20pt}}c@{}}
\includegraphics[width=.22\linewidth]{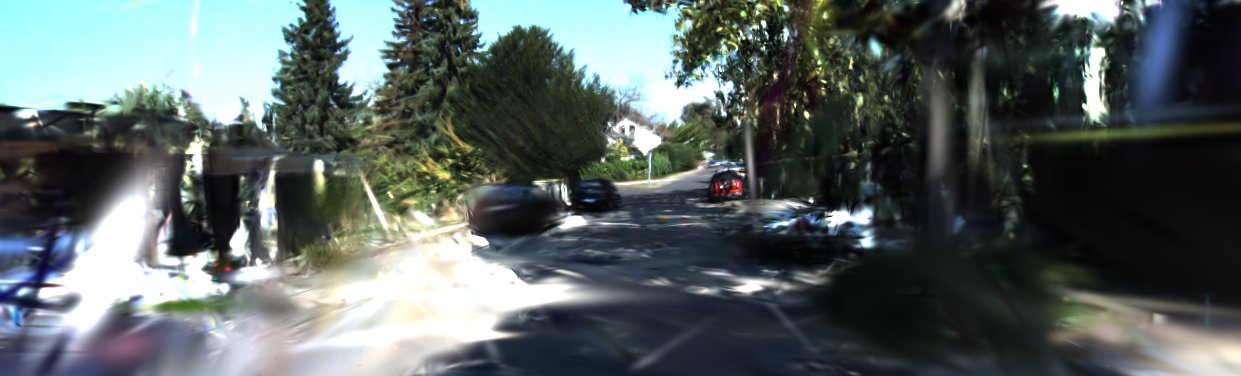} & &
\includegraphics[width=.22\linewidth]{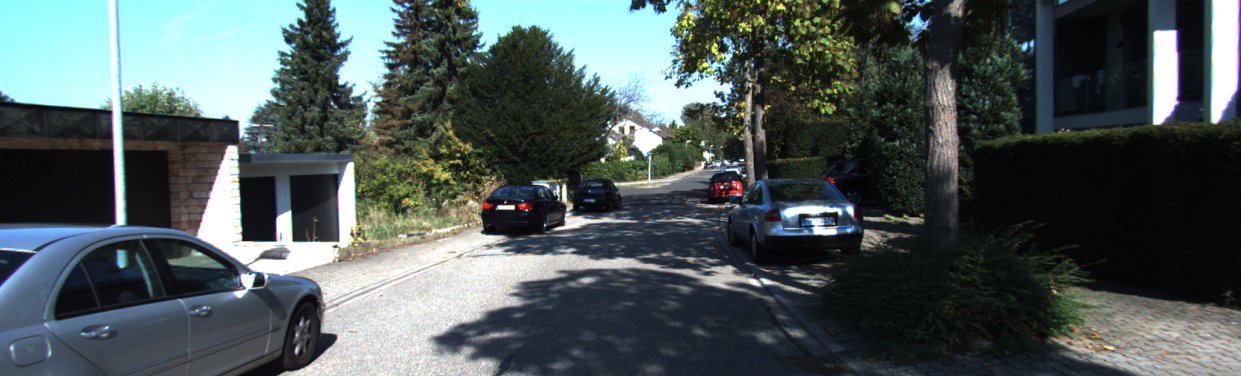} \\
$\lambda{=}0$ (no depth prior) & & Ground Truth \\
\end{tabular}

\caption{Ablation on $\lambda$ and $\tau$ for Splatfacto. Depth supervision improves thin structures (street pole) and reduces boundary artifacts vs.\ the RGB-only baseline ($\lambda{=}0$).}
\label{fig:ablation_splatfacto}
\end{figure*}

The RGB-only Mip-NeRF-360 baseline achieves 20.378 PSNR, 0.601 SSIM, 0.409 LPIPS, 0.0857 AbsRel and 2.703 RMSE. The best PSNR is achieved with $\tau{=}1.0$, $\lambda{=}0.15$: 20.607 PSNR. \Rhi{fxQT}{Because $\tau{=}1.0$ represents the unmasked condition, the optimal configuration bypasses the mask entirely. The best masked setting ($\tau{=}0.16$, $\lambda{=}0.15$: 20.581 PSNR) trails the global peak by 0.026\,dB. All depth-supervised settings, including global, degrade SSIM, LPIPS, and geometry relative to RGB-only (RMSE rises from 2.703 to ${\geq}3.532$). Multi-seed evaluation ($20.57{\pm}0.04$\,dB PSNR) confirms this gap falls within run-to-run variance.}

\begin{figure}[t]
\centering 
\begin{subfigure}[b]{0.48\linewidth} 
    \centering 
    \includegraphics[width=\linewidth]{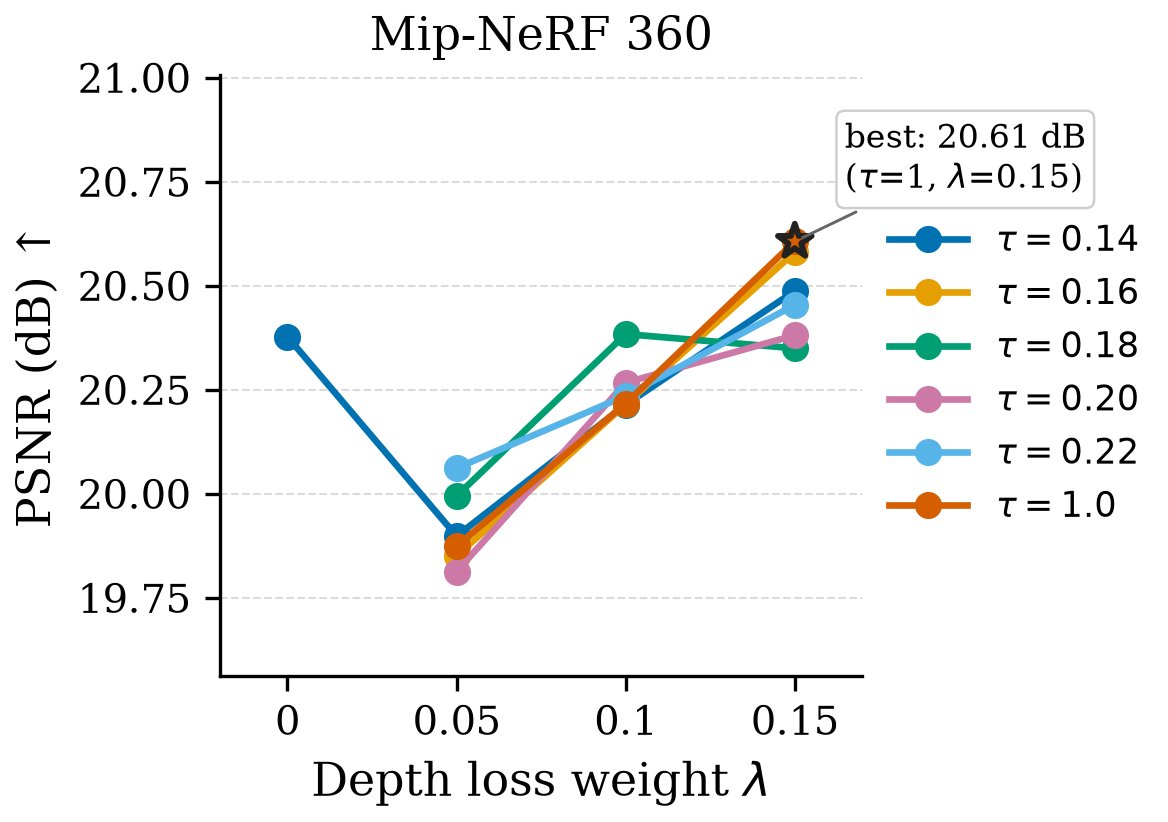} 
    \caption{PSNR increases monotonically with $\lambda$ across nearly all $\tau$ values, showing consistent benefit from stronger depth regularization.} 
    \label{fig:left_image} 
\end{subfigure}%
\hfill 
\begin{subfigure}[b]{0.48\linewidth} 
    \centering 
    \includegraphics[width=\linewidth]{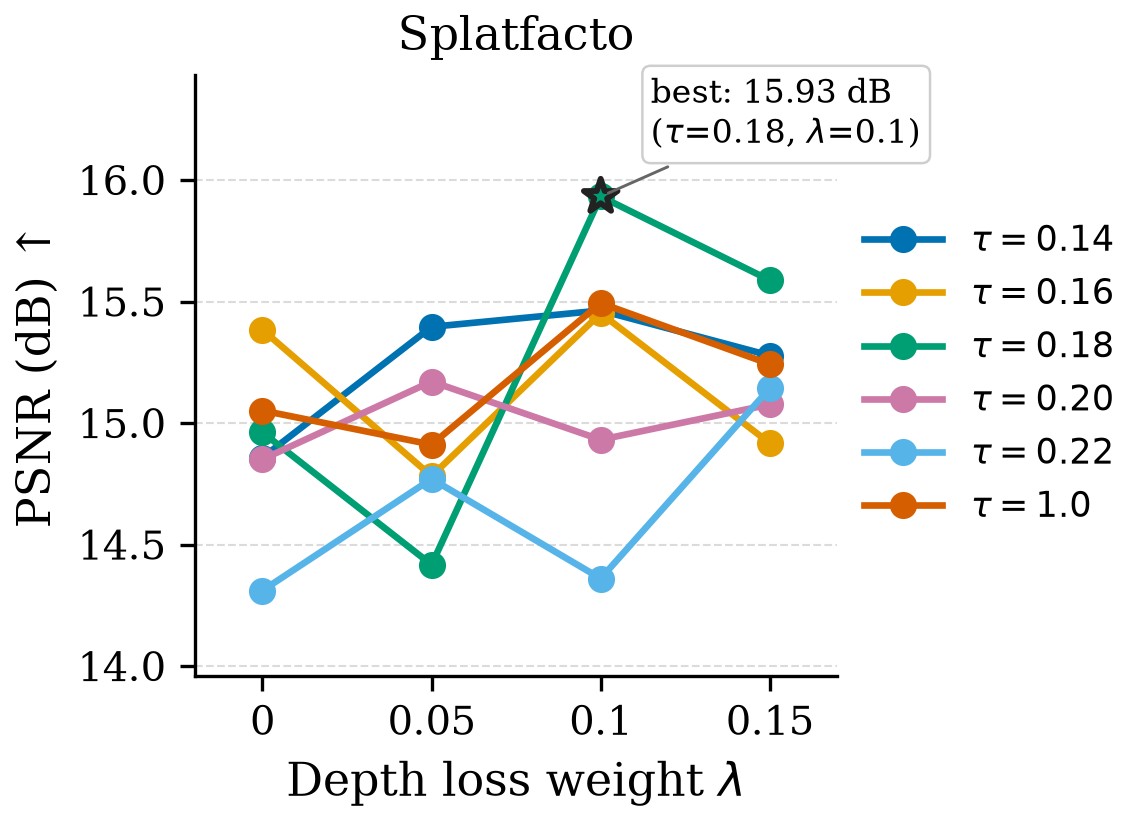} 
    \caption{PSNR peaks sharply at $\tau{=}0.18$, $\lambda{=}0.1$ (15.93 dB), showing sensitivity to depth supervision and threshold values.} 
    \label{fig:right_image} 
\end{subfigure} 
\caption{PSNR as a function of depth loss weight $\lambda$ on KITTI Seq02, evaluated across six threshold values $\tau \in \{0.14, 0.16, 0.18, 0.20, 0.22, 1.0\}$.} 
\label{fig:combined_main_figure} 
\end{figure}

\begin{table}[t]
\centering
\small
\resizebox{\linewidth}{!}{
\begin{tabular}{c c c c c c}
\hline
$\tau$ & $\lambda$ & PSNR$\uparrow$ & SSIM$\uparrow$ & LPIPS$\downarrow$ & AbsRel$\downarrow$ / RMSE$\downarrow$ \\
\hline
RGB-only & 0 & 20.378 & \textbf{0.601} & \textbf{0.409} & \textbf{0.0857} / \textbf{2.703} \\
0.14 & 0.15 & 20.488 & 0.596 & 0.413 & 0.1113 / 3.663 \\
0.16 & 0.15 & 20.581 & 0.596 & 0.415 & 0.1098 / 3.654 \\
0.18 & 0.10 & 20.384 & 0.594 & 0.416 & 0.1044 / 3.532 \\
0.20 & 0.15 & 20.381 & 0.594 & 0.417 & 0.1097 / 3.658 \\
0.22 & 0.15 & 20.454 & 0.594 & 0.416 & 0.1088 / 3.603 \\
1.00 \Rhi{fxQT}{(global)} & 0.15 & \textbf{20.607} & 0.595 & 0.412 & 0.1074 / 3.580 \\
\hline
\end{tabular}
}
\caption{
Compact summary of Mip-NeRF-360 results on KITTISeq02 under the every2 sparse-view setting. We show the RGB-only baseline and the strongest representative setting for each threshold. Since $\lambda{=}0$ disables the depth loss, the RGB-only result is shared across thresholds. \Rhi{fxQT}{No masked setting surpasses the global ($\tau{=}1.00$) row.}
}
\label{tab:mipnerf_results_compact}
\end{table}

We further evaluate representative Mip-NeRF-360 settings on KITTISeq05. As shown in Table~\ref{tab:mipnerf_seq05_results}, both global depth supervision and low-error masked supervision reduce rendering quality and increase depth error compared with RGB-only training. \Rhi{fxQT}{The mask recovers 0.223\,dB PSNR over global (16.612 vs.\ 16.389), so it mitigates rather than reverses the harm.} This supports the observation that monocular depth priors are not reliably beneficial for the implicit density representation.

\begin{table}[t]
\centering
\small
\resizebox{\linewidth}{!}{
\begin{tabular}{c c c c c c}
\hline
$\tau$ & $\lambda$ & PSNR$\uparrow$ & SSIM$\uparrow$ & LPIPS$\downarrow$ & AbsRel$\downarrow$ / RMSE$\downarrow$ \\
\hline
RGB-only & 0 & \textbf{17.068} & \textbf{0.546} & \textbf{0.529} & \textbf{0.1166} / \textbf{2.978} \\
1.00 \Rhi{fxQT}{(global)} & 0.15 & 16.389 & 0.527 & 0.569 & 0.1527 / 4.803 \\
0.18 & 0.15 & 16.612 & 0.530 & 0.563 & 0.1399 / 4.454 \\
\hline
\end{tabular}
}
\caption{
Additional Mip-NeRF-360 results on KITTISeq05. Global and masked DA-V2 supervision both degrade rendering and geometry relative to RGB-only training, although the low-error mask reduces the degradation compared with global supervision.
}
\label{tab:mipnerf_seq05_results}
\end{table}

\subsection{Splatfacto Results}

\noindent\textbf{Takeaway.} Splatfacto benefits more clearly from masked monocular depth supervision in sparse forward-facing KITTI scenes. Matched-ratio ablations show that the low-error mask outperforms high-error and random masks, indicating that the gain comes from selecting more reliable pixels rather than simply reducing the number of depth-supervised pixels. \Rhi{fxQT}{Against global (unmasked) supervision, PSNR gains of 0.44--0.70\,dB are observed across the three KITTI fragments at tied or better RMSE: the mask's contribution is rendering fidelity, not geometry.}

Table~\ref{tab:splatfacto_results_compact} summarizes Splatfacto results on KITTISeq02~\cite{kitti,3dgs,nerfstudio}. The RGB-only baseline achieves 14.903 PSNR, 0.433 SSIM, 0.446 LPIPS, and 0.542 RMSE. The best result is $\tau=0.18$, $\lambda=0.10$: 15.932 PSNR, 0.477 SSIM, 0.408 LPIPS, and 0.100 RMSE. \Rhi{fxQT}{Against global supervision ($\tau{=}1.0$, $\lambda{=}0.10$: 15.494 PSNR, 0.101 RMSE), a PSNR gain of 0.44\,dB is observed at tied or better RMSE: the RMSE drop from 0.542 comes from using a depth prior at all, and masking is what improves rendering fidelity. Over three seeds: masked $15.30{\pm}0.57$ vs.\ global $15.07{\pm}0.38$ (mean $+0.23$\,dB); the $-80\%$ RMSE improvement over RGB-only is consistent across seeds.}

At $\tau{=}0.18$, increasing $\lambda$ from $0.10$ to $0.15$ improves RMSE from 0.100 to 0.096 but decreases PSNR from 15.932 to 15.588 and increases LPIPS from 0.408 to 0.436, revealing a rendering--geometry tradeoff.

\begin{table}[t]
\centering
\small
\resizebox{\linewidth}{!}{
\begin{tabular}{c c c c c c}
\hline
$\tau$ & $\lambda$ & PSNR$\uparrow$ & SSIM$\uparrow$ & LPIPS$\downarrow$ & RMSE$\downarrow$ \\
\hline
RGB-only & 0 & 14.903 & 0.433 & 0.446 & 0.542 \\
0.14 & 0.10 & 15.463 & 0.454 & 0.436 & 0.105 \\
0.16 & 0.10 & 15.452 & 0.454 & 0.434 & 0.106 \\
0.18 & 0.10 & \textbf{15.932} & \textbf{0.477} & \textbf{0.408} & 0.100 \\
0.18 & 0.15 & 15.588 & 0.459 & 0.436 & \textbf{0.096} \\
0.20 & 0.05 & 15.171 & 0.431 & 0.463 & 0.129 \\
0.22 & 0.15 & 15.143 & 0.443 & 0.452 & 0.103 \\
1.00 \Rhi{fxQT}{(global)} & 0.10 & 15.494 & 0.448 & 0.434 & 0.101 \\
\hline
\end{tabular}
}
\caption{
Compact summary of Splatfacto results on KITTISeq02 under the every2 sparse-view setting. The table reports the RGB-only baseline and the strongest representative settings from the hyperparameter sweep. At $\tau{=}0.18$, we include both $\lambda{=}0.10$ and $\lambda{=}0.15$ because the former gives the best rendering quality while the latter gives the best RMSE. \Rhi{fxQT}{Relative to the global ($\tau{=}1.00$) row, a PSNR gain of 0.44\,dB is observed at tied or better RMSE.}
}

\label{tab:splatfacto_results_compact}
\end{table}

Table~\ref{tab:splatfacto_mask_ablation} shows a matched-ratio ablation at $\tau{=}0.18$, $\lambda{=}0.10$: the low-error mask achieves the best PSNR, SSIM, LPIPS, and RMSE, confirming that region selection matters, not just pixel count. 

\begin{table}[t]
\centering\small
\setlength{\tabcolsep}{5pt}
\begin{tabular}{l c c c c}
\toprule
Mask & PSNR$\uparrow$ & SSIM$\uparrow$ & LPIPS$\downarrow$ & RMSE$\downarrow$ \\
\midrule
High-error   & 14.932 & 0.437 & 0.455 & 0.111 \\[2pt]
Random       & \shortstack{15.036\\[-1pt]{\footnotesize$\pm$0.341}}
             & \shortstack{0.442\\[-1pt]{\footnotesize$\pm$0.013}}
             & \shortstack{0.456\\[-1pt]{\footnotesize$\pm$0.017}}
             & \shortstack{0.109\\[-1pt]{\footnotesize$\pm$0.002}} \\[6pt]
Low-error (ours) & \textbf{15.932} & \textbf{0.477} & \textbf{0.408} & \textbf{0.100} \\
\bottomrule
\end{tabular}
\caption{Matched-ratio mask ablation for Splatfacto on KITTISeq02 ($\tau{=}0.18$, $\lambda{=}0.10$). All three masks supervise the same pixel count. Random: mean$\pm$std over 3 seeds. \textbf{Bold}: best.}
\label{tab:splatfacto_mask_ablation}
\end{table}

\Rhi{fxQT}{Table~\ref{tab:competing_masks} places these results next to existing depth-supervised methods on KITTISeq02, all retrained for 50k iterations on our sparse split. Unmasked DA-V2 supervision already improves on RGB-only ($+0.59$\,dB), and masking yields a further $+0.44$\,dB. DNGaussian, DepthRegGS and SparseGS fall well below our monocular result despite using GT LiDAR, as they target indoor or object-centric scenes. Only DN-Splatter leads ($+0.29$\,dB PSNR, $-0.12$ LPIPS), using GT LiDAR unavailable at test time, so its margin reflects depth-source quality rather than masking.}

\begin{table}[t]\centering\small
\resizebox{\linewidth}{!}{%
\begin{tabular}{lccc}
\toprule
Method & PSNR$\uparrow$ & SSIM$\uparrow$ & LPIPS$\downarrow$ \\
\midrule
RGB-only (no depth, $\lambda{=}0$)                  & 14.903 & 0.433 & 0.446 \\
DA-V2 depth, no mask ($\lambda{=}0.1$)              & 15.494 & 0.448 & 0.434 \\
\textbf{Ours} ($\tau{=}0.18$, $\lambda{=}0.1$)      & 15.932 & 0.477 & 0.408 \\
\midrule
\Rhi{fxQT}{DNGaussian~\cite{dngaussian} (GT LiDAR)}                         & \Rhi{fxQT}{9.98}  & \Rhi{fxQT}{0.303} & \Rhi{fxQT}{0.710} \\
\Rhi{fxQT}{DepthRegGS~\cite{depth_regularized_3dgs} (GT LiDAR)}             & \Rhi{fxQT}{8.71}  & \Rhi{fxQT}{0.229} & \Rhi{fxQT}{0.737} \\
\Rhi{fxQT}{SparseGS~\cite{sparsegs} (GT LiDAR)}                             & \Rhi{fxQT}{12.20} & \Rhi{fxQT}{0.359} & \Rhi{fxQT}{0.648} \\
DN-Splatter~\cite{dn_splatter} (GT LiDAR, $\lambda{=}0.2$) & \textbf{16.22} & \textbf{0.489} & \textbf{0.289} \\
\bottomrule
\end{tabular}}
\caption{\Rhi{fxQT}{Comparison to depth-supervised methods on KITTISeq02 Frag.~034 (50k iterations). The top group uses no GT depth and repeats the RGB-only, $\tau{=}1.00$ and $\tau{=}0.18$ rows of Table~\ref{tab:splatfacto_results_compact}; the bottom group uses GT LiDAR. DNGaussian, DepthRegGS, and SparseGS target indoor or object-centric scenes. \textbf{Bold}: best per column.}}
\label{tab:competing_masks}
\end{table}

\Rhi{fxQT}{To show the effect generalizes beyond KITTISeq02, Table~\ref{tab:splatfacto_multi_seq} aggregates three KITTI fragments (02/034, 05/018, 00/027). Masked ($\tau{=}0.18$) outperforms global on every metric across all three sequences, with PSNR gains of $+0.44$, $+0.64$ and $+0.70$\,dB at tied or better RMSE.}

\begin{table}[t]
\centering
\small
\resizebox{\linewidth}{!}{
\begin{tabular}{c l c c c c}
\toprule
Seq & Method & PSNR$\uparrow$ & SSIM$\uparrow$ & LPIPS$\downarrow$ & RMSE$\downarrow$ \\
\midrule
\multirow{3}{*}{02/034} & RGB-only & 14.90 & 0.433 & 0.446 & 0.542 \\
& Global ($\tau{=}1.0$) & 15.49 & 0.448 & 0.434 & 0.101 \\
& Masked ($\tau{=}0.18$) & \textbf{15.93} & \textbf{0.477} & \textbf{0.408} & \textbf{0.100} \\
\cmidrule{2-6}
\multirow{3}{*}{05/018} & RGB-only & 14.89 & 0.521 & 0.493 & 0.807 \\
& Global ($\tau{=}1.0$) & 15.26 & 0.534 & 0.469 & 0.096 \\
& Masked ($\tau{=}0.18$) & \textbf{15.90} & \textbf{0.548} & \textbf{0.446} & \textbf{0.096} \\
\cmidrule{2-6}
\multirow{3}{*}{00/027} & RGB-only & 14.67 & 0.487 & 0.401 & 0.514 \\
& Global ($\tau{=}1.0$) & 16.69 & 0.554 & 0.302 & 0.125 \\
& Masked ($\tau{=}0.18$) & \textbf{17.39} & \textbf{0.571} & \textbf{0.288} & \textbf{0.114} \\
\bottomrule
\end{tabular}
}
\caption{\Rhi{fxQT}{Splatfacto results across three KITTI sequences. \textbf{Bold}: best per sequence.}}
\label{tab:splatfacto_multi_seq}
\end{table}

\begin{figure*}[!htbp]
\centering
\setlength{\tabcolsep}{1.5pt}
\renewcommand{\arraystretch}{0.3}
\footnotesize

\begin{tabular}{@{}c@{\hspace{4pt}}cccccc@{}}
 & $\tau{=}0.14$ & $\tau{=}0.16$ & $\tau{=}0.18$ & $\tau{=}0.20$ & $\tau{=}0.22$ & global \\

\rotatebox{90}{$\lambda{=}0.05$} &
\includegraphics[width=.115\linewidth]{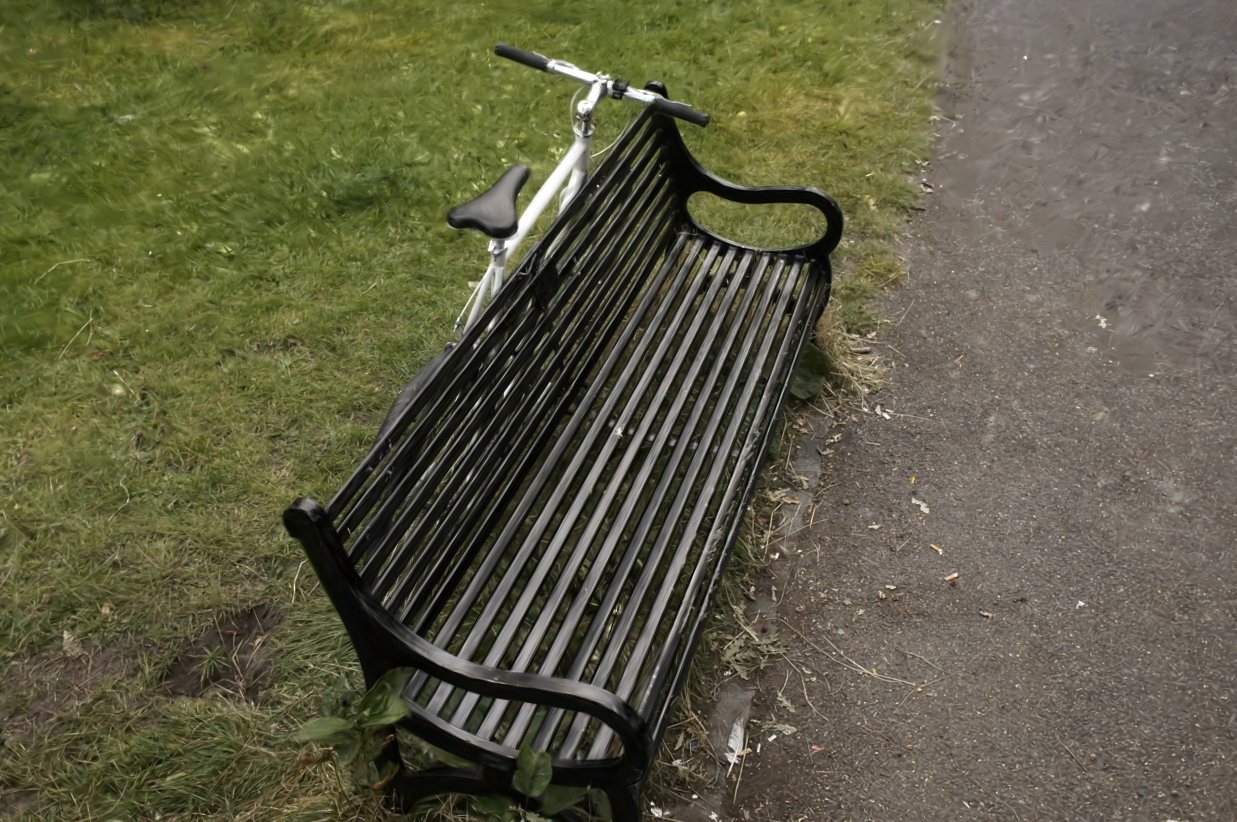} &
\includegraphics[width=.115\linewidth]{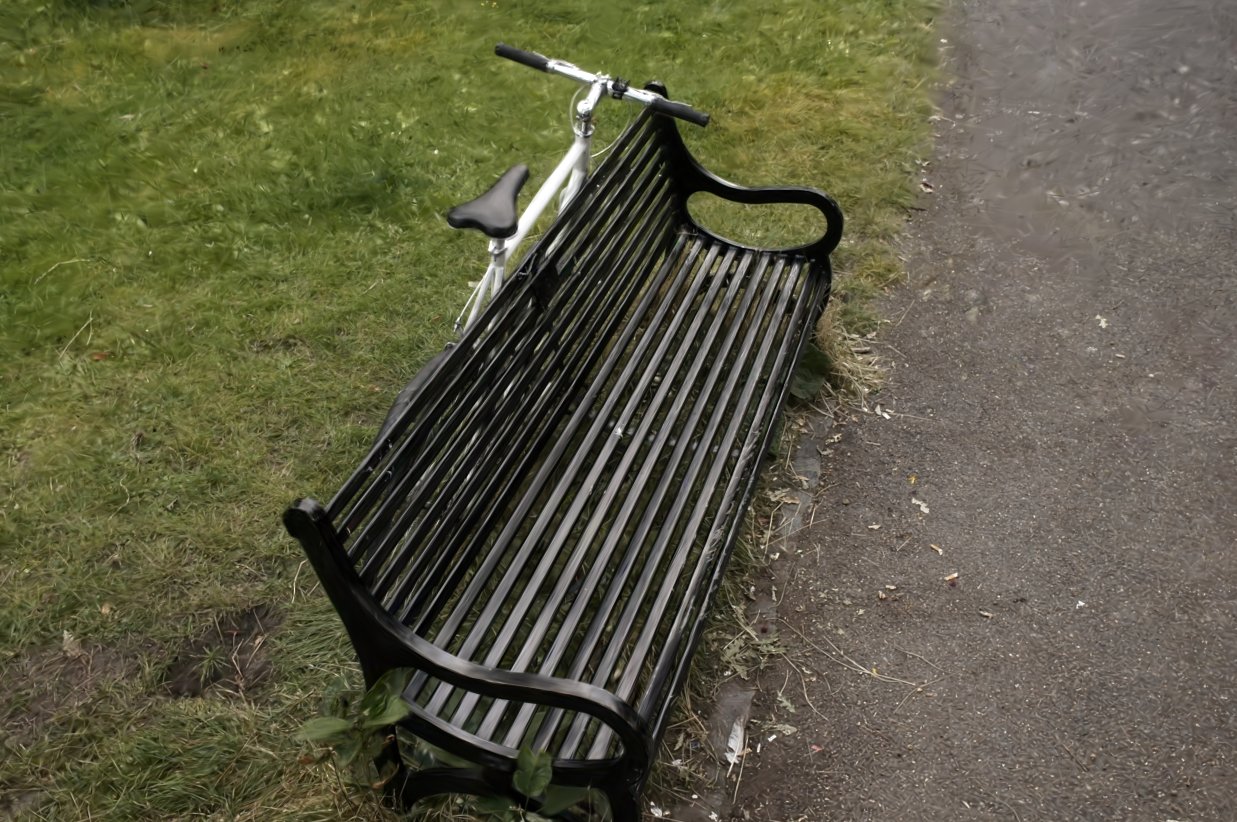} &
\includegraphics[width=.115\linewidth]{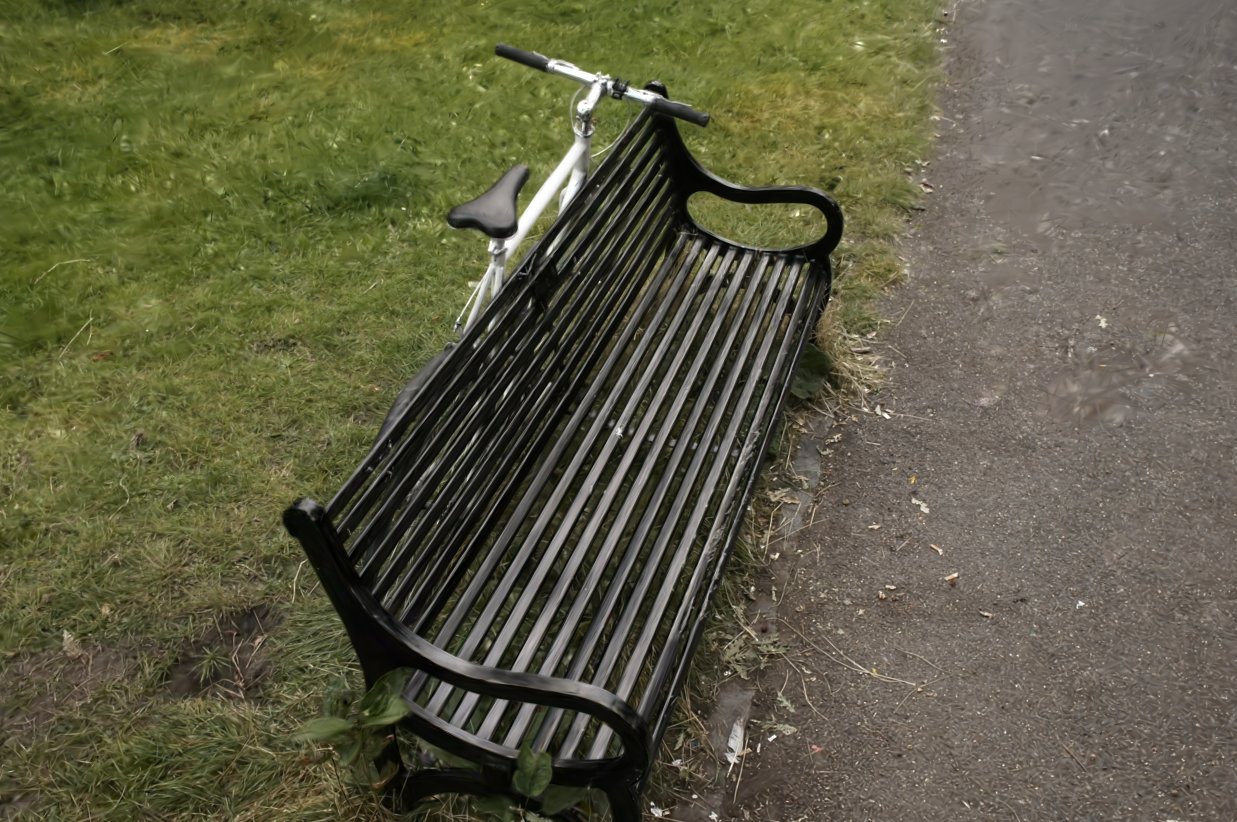} &
\includegraphics[width=.115\linewidth]{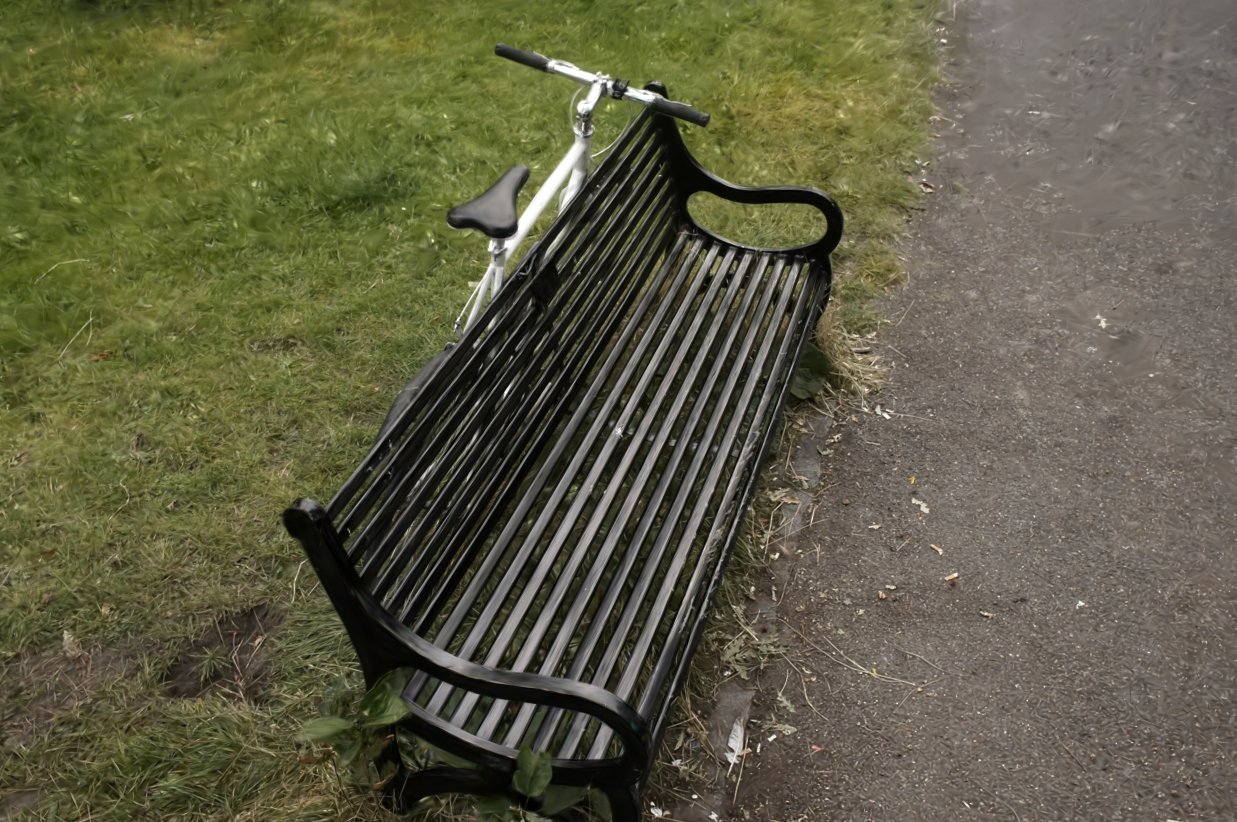} &
\includegraphics[width=.115\linewidth]{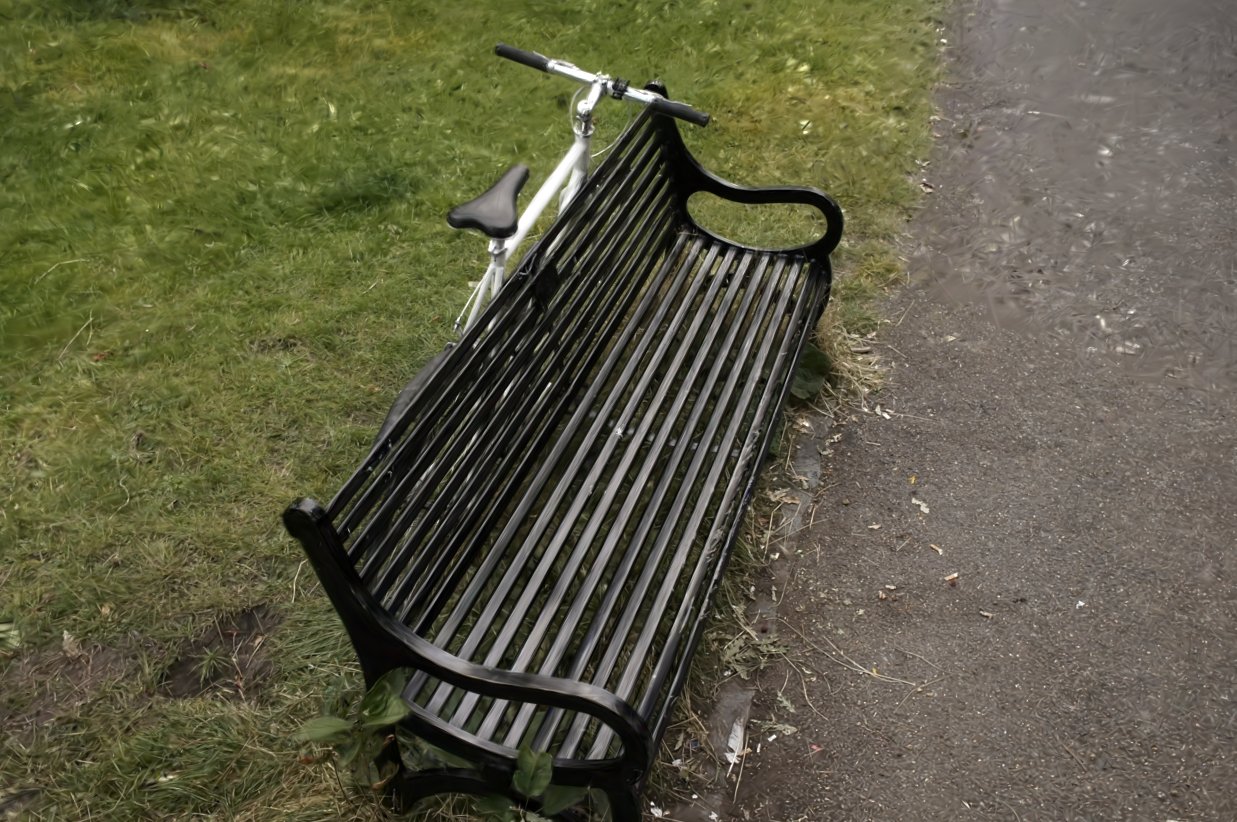} &
\includegraphics[width=.115\linewidth]{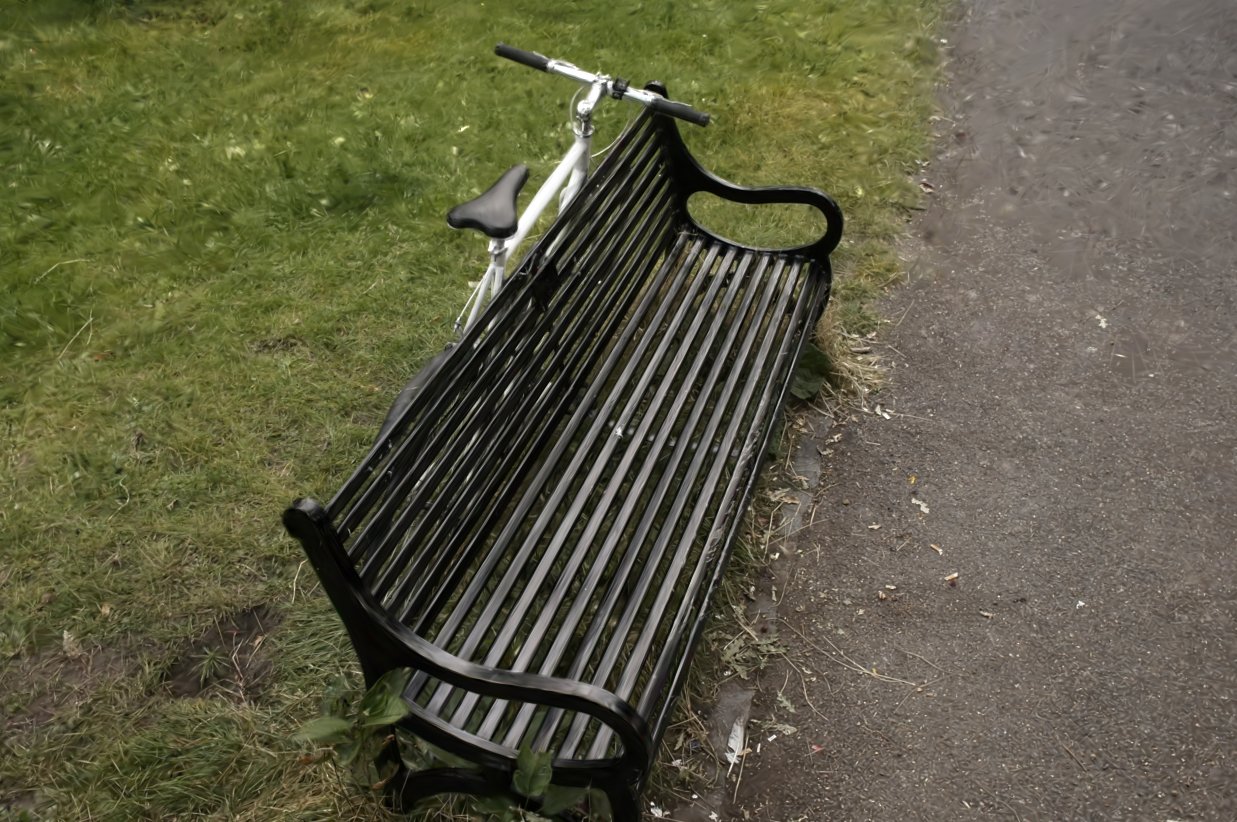} \\

\rotatebox{90}{$\lambda{=}0.10$} &
\includegraphics[width=.115\linewidth]{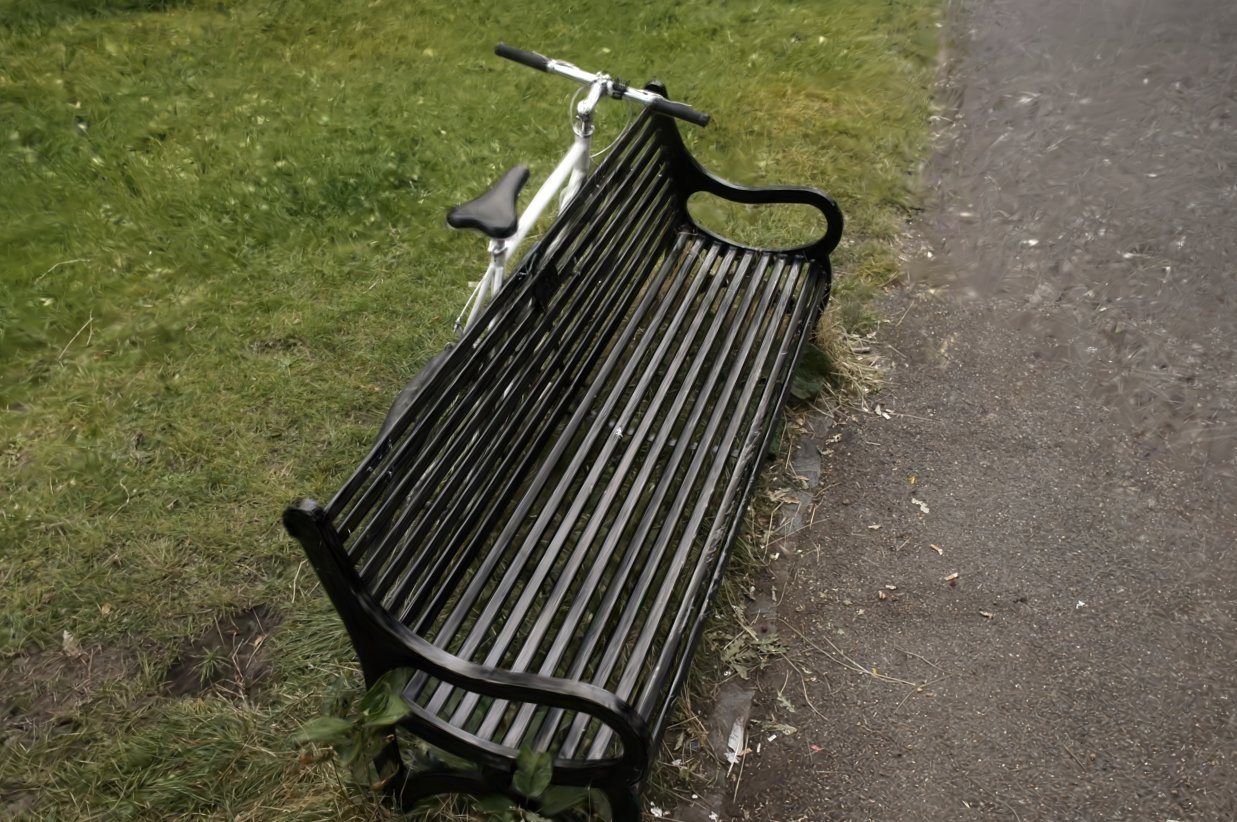} &
\includegraphics[width=.115\linewidth]{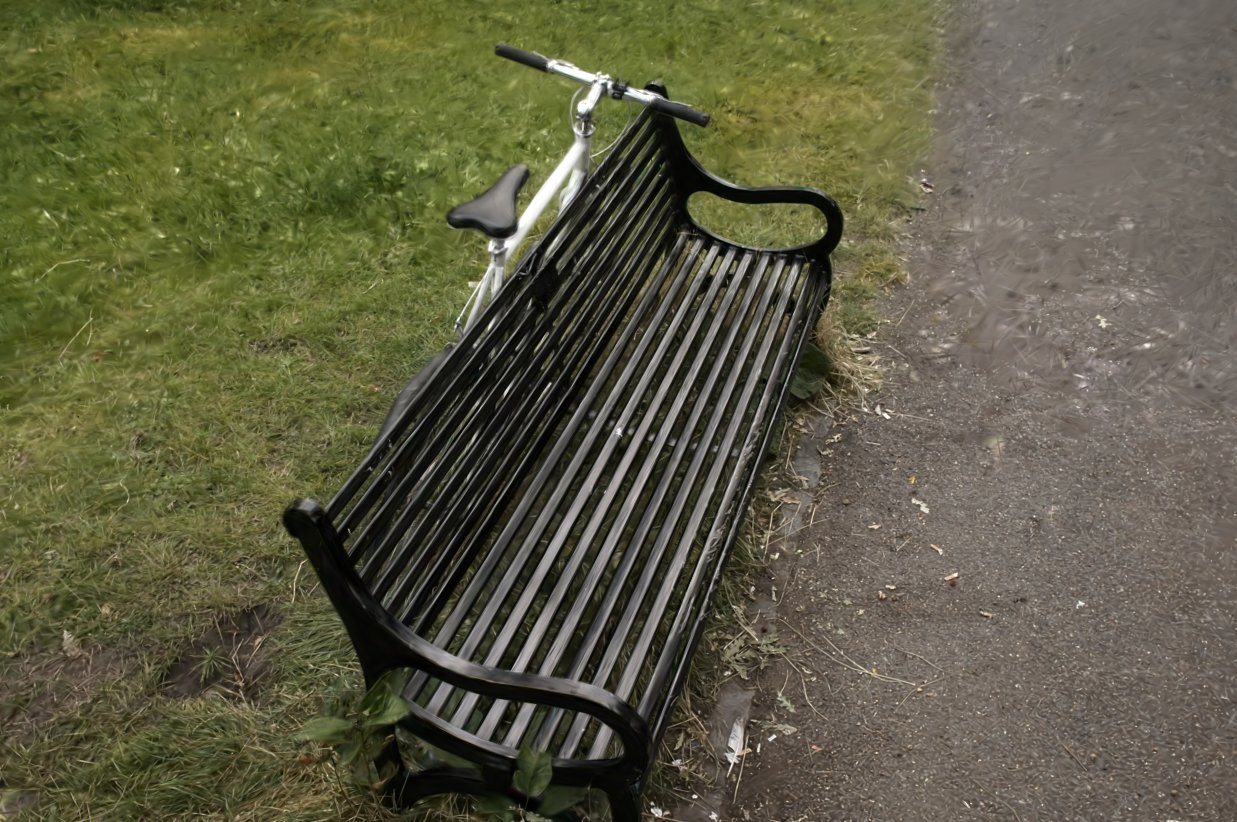} &
\includegraphics[width=.115\linewidth]{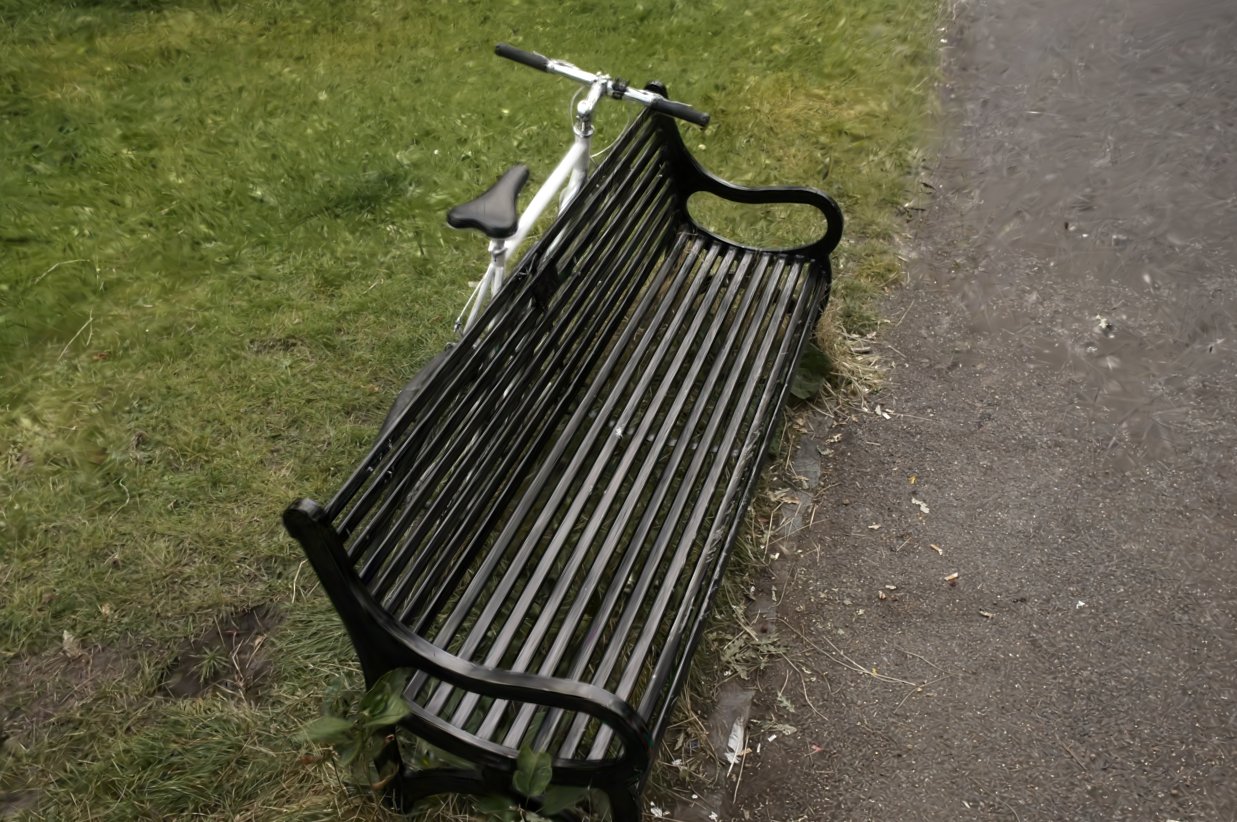} &
\includegraphics[width=.115\linewidth]{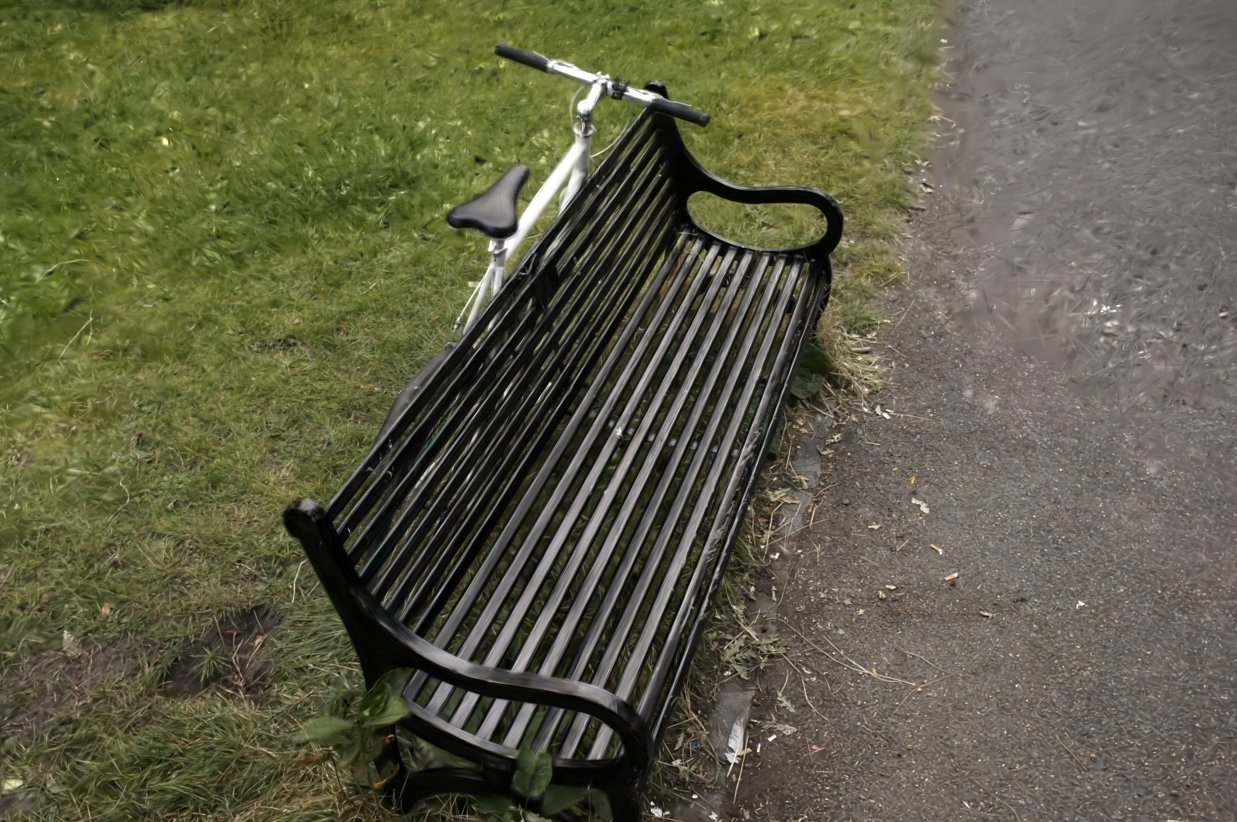} &
\includegraphics[width=.115\linewidth]{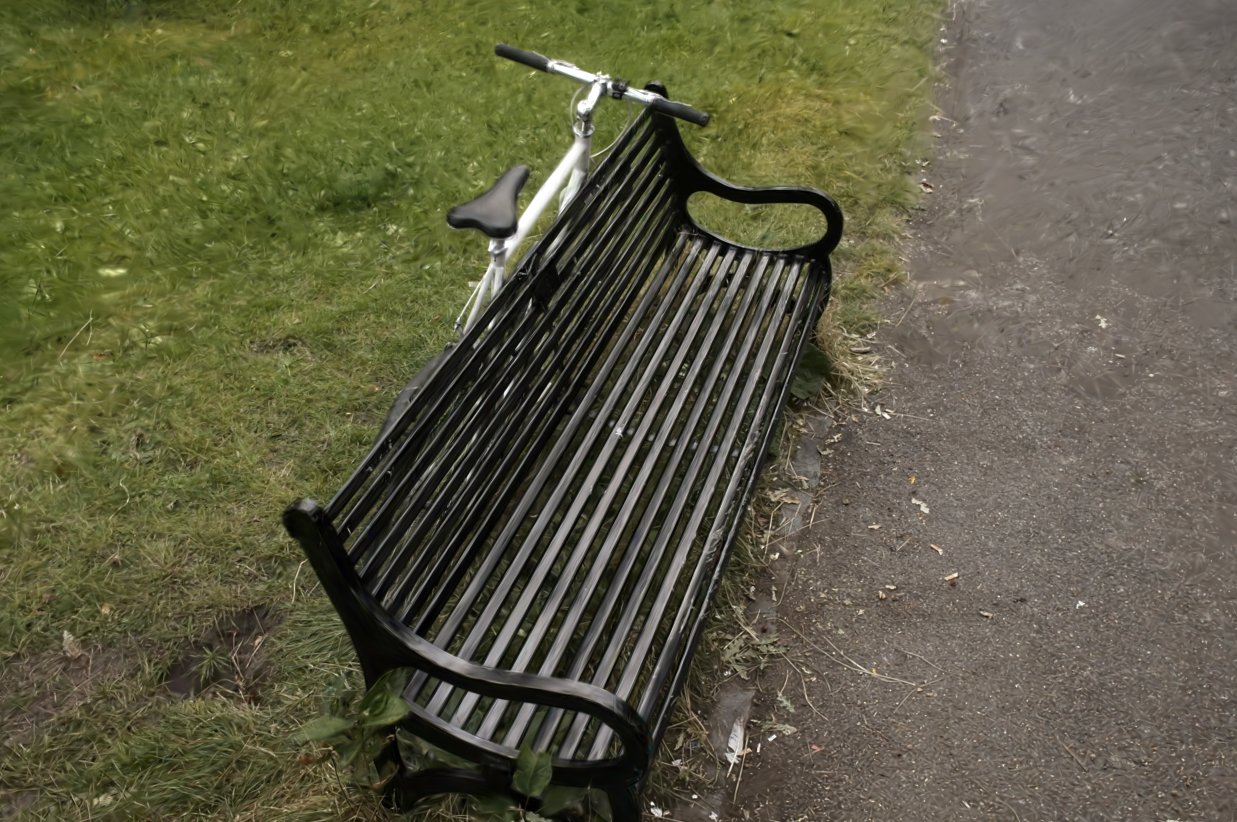} &
\includegraphics[width=.115\linewidth]{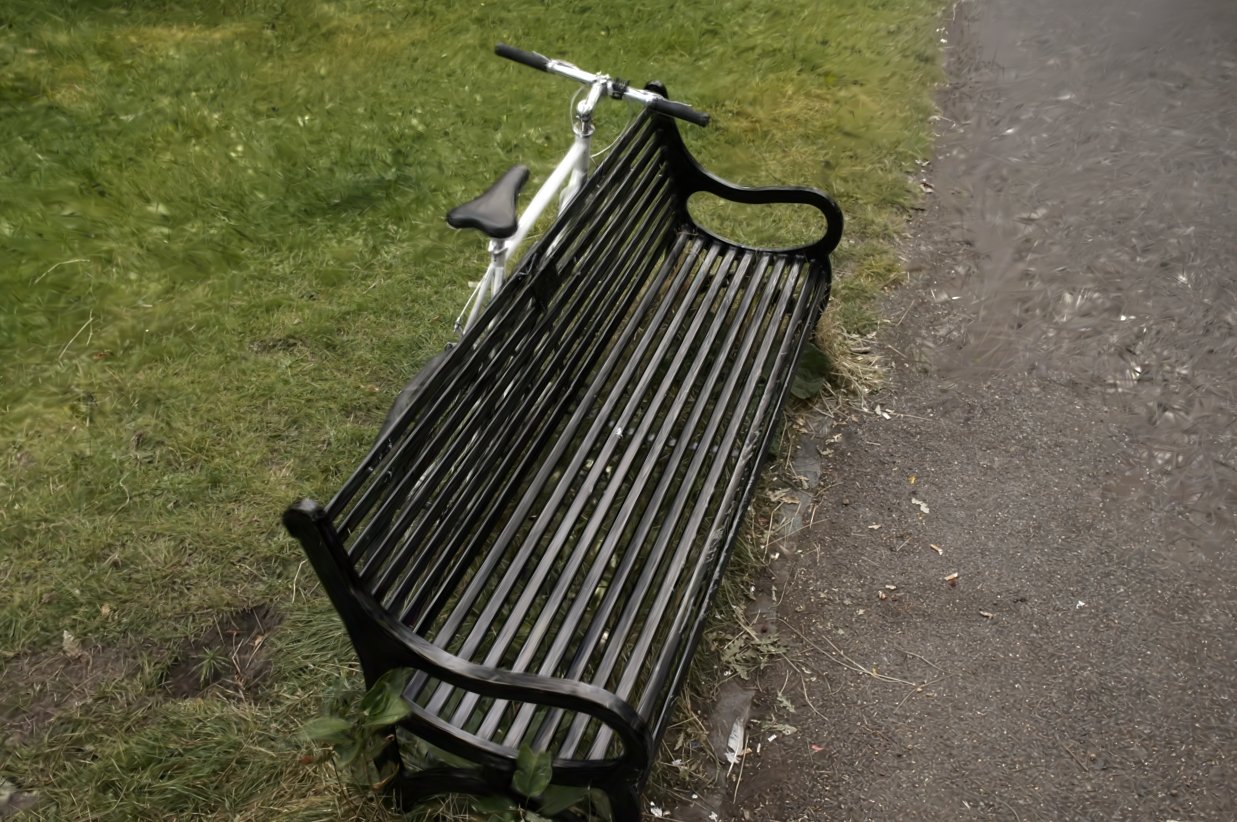} \\

\rotatebox{90}{$\lambda{=}0.15$} &
\includegraphics[width=.115\linewidth]{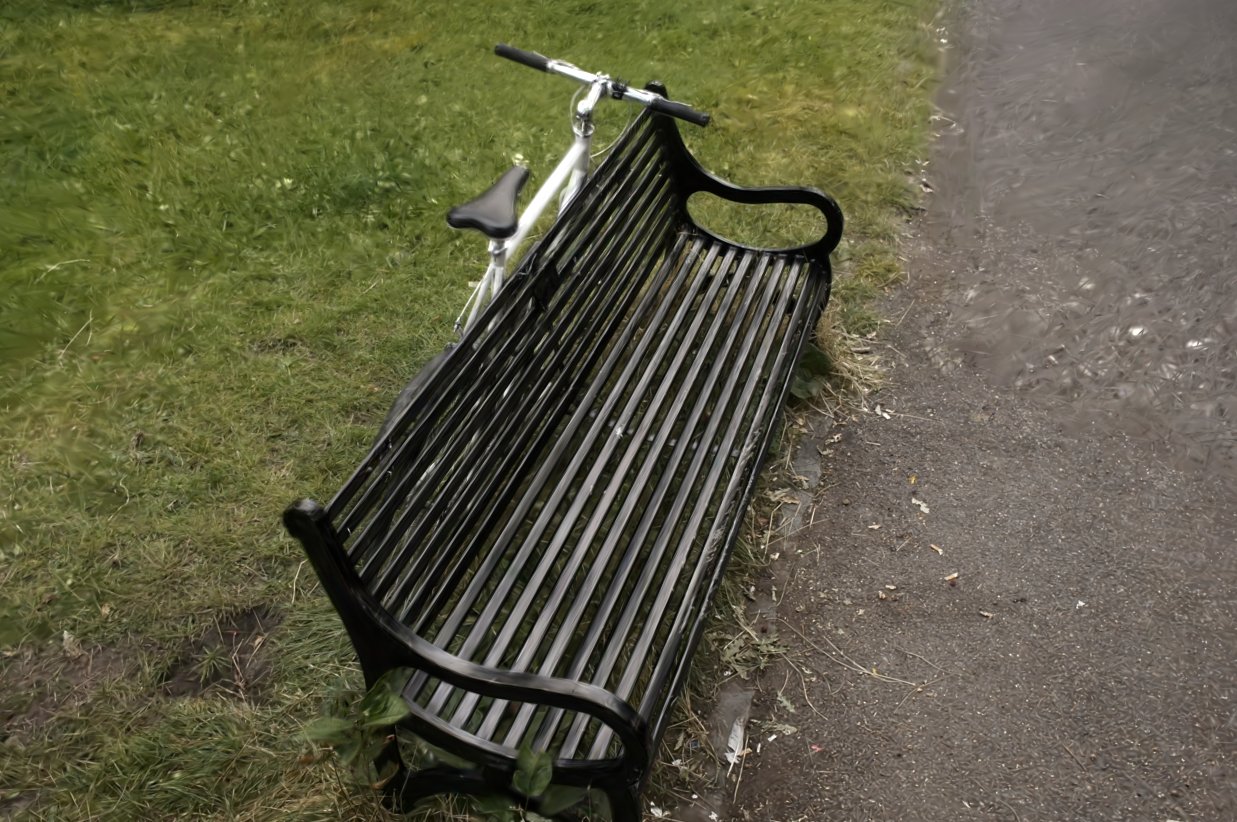} &
\includegraphics[width=.115\linewidth]{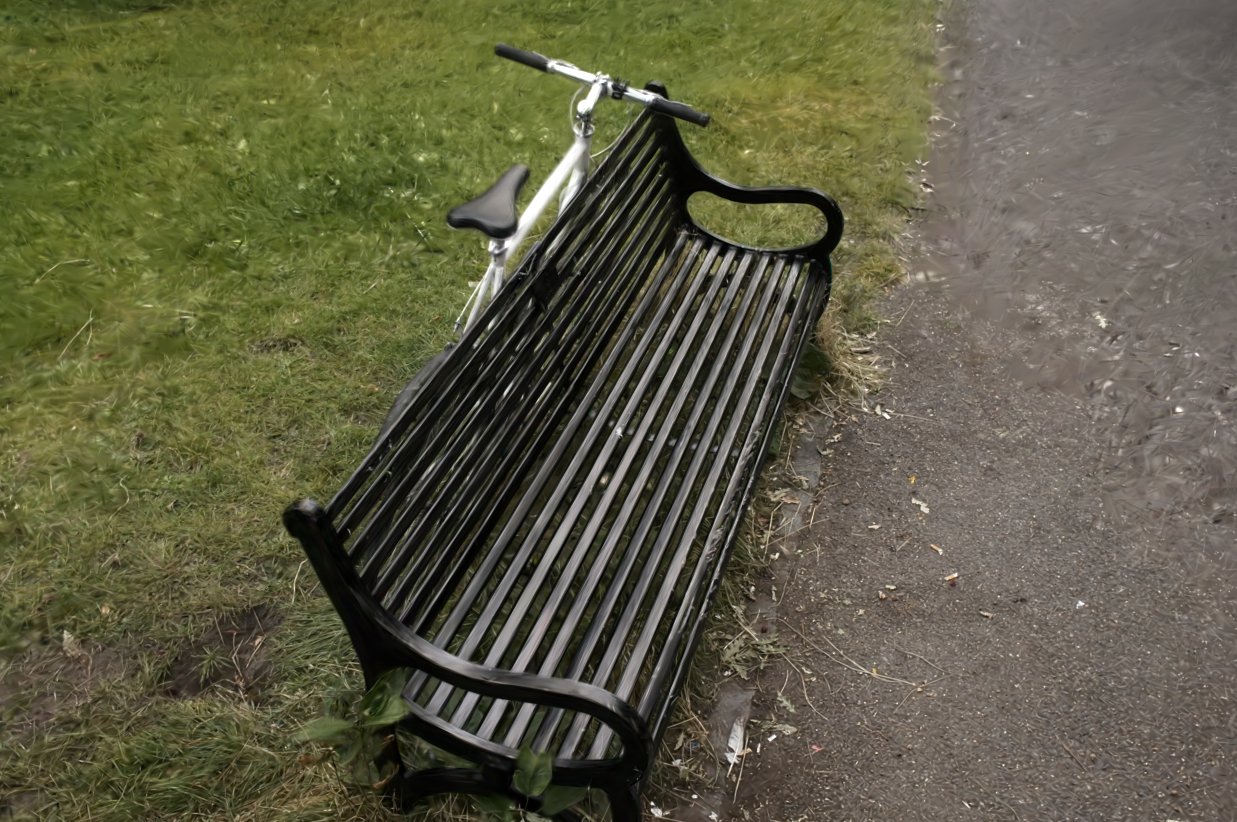} &
\includegraphics[width=.115\linewidth]{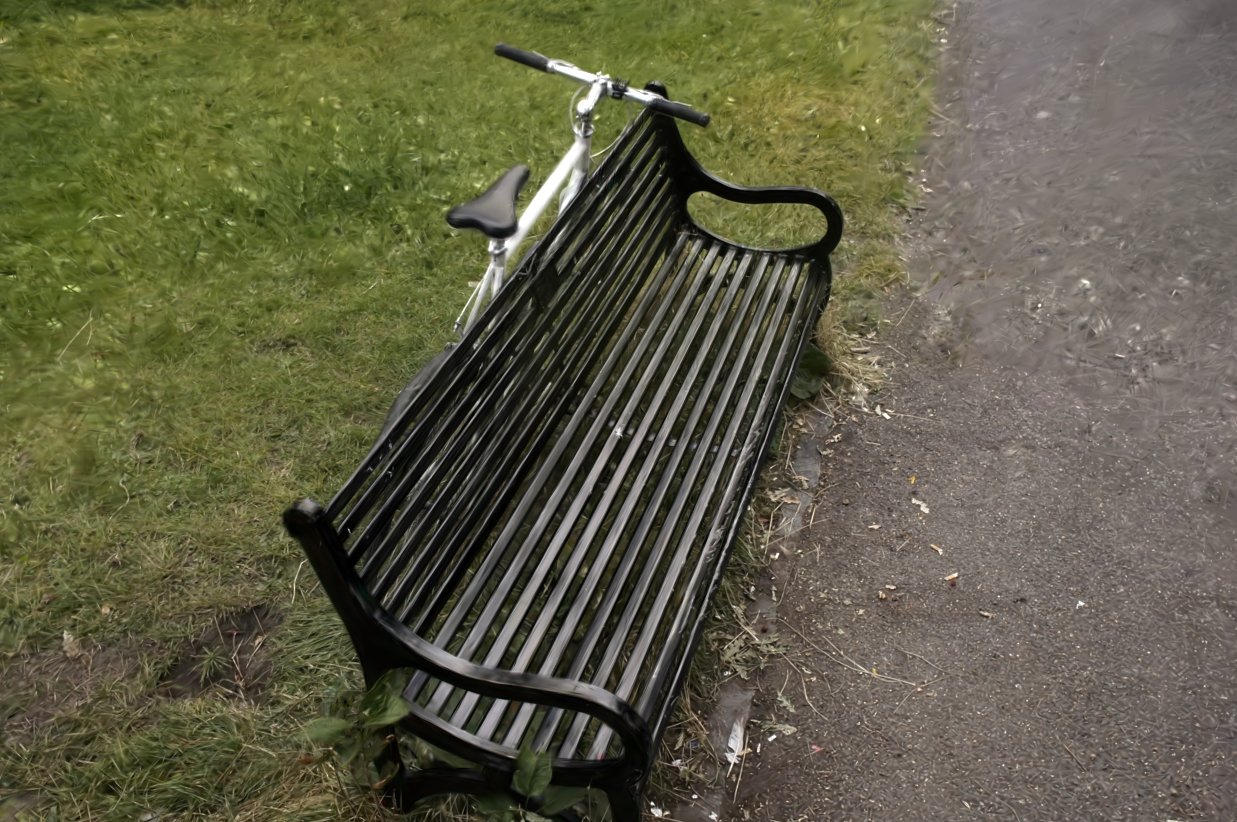} &
\includegraphics[width=.115\linewidth]{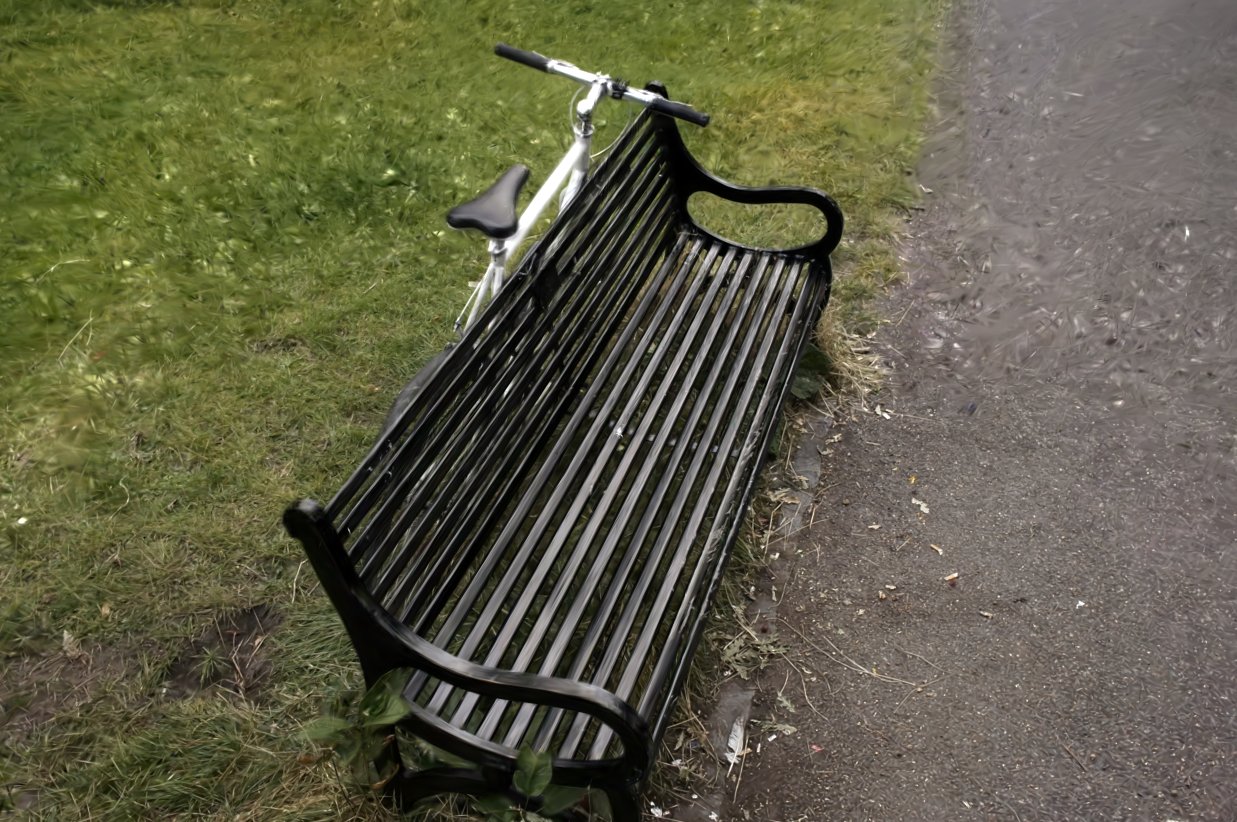} &
\includegraphics[width=.115\linewidth]{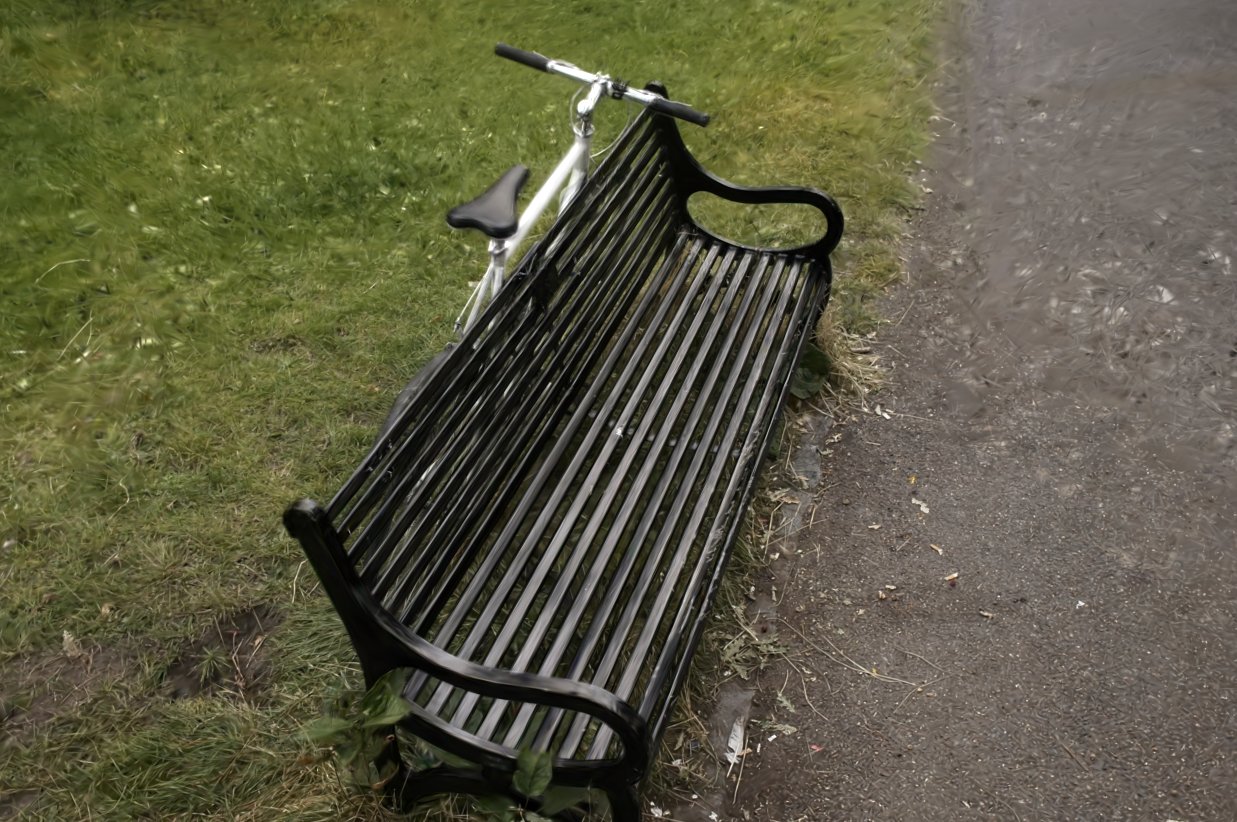} &
\includegraphics[width=.115\linewidth]{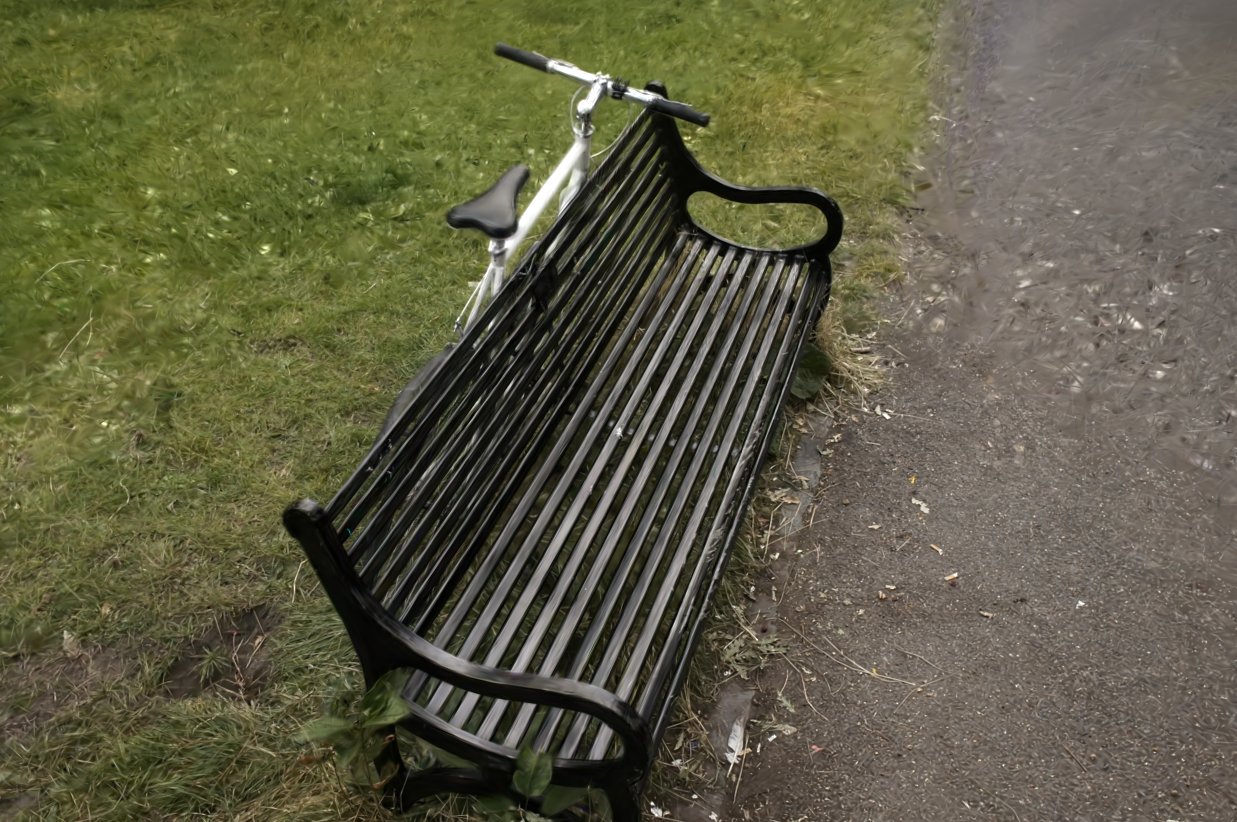} \\
\end{tabular}

\vspace{6pt}

\begin{tabular}{@{}cc@{\hspace{20pt}}c@{}}
\includegraphics[width=.15\linewidth]{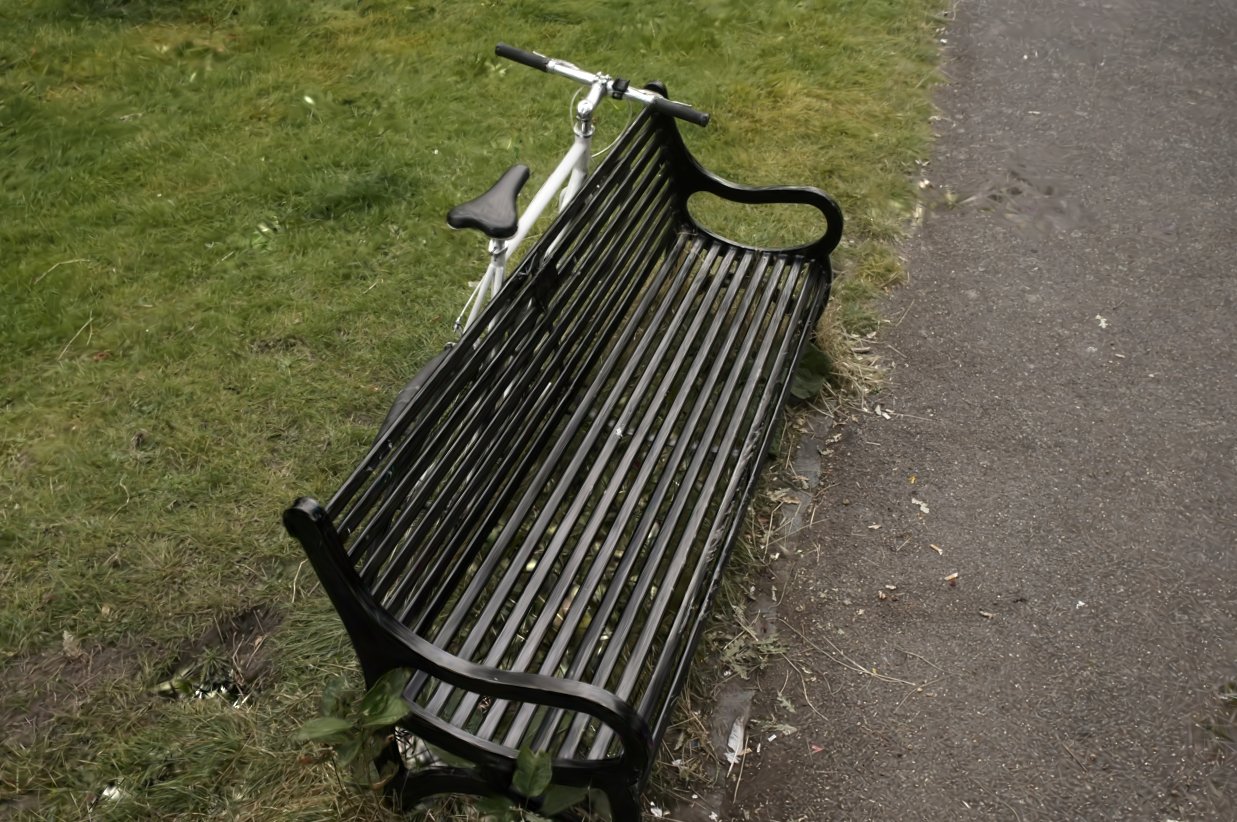} & &
\includegraphics[width=.15\linewidth]{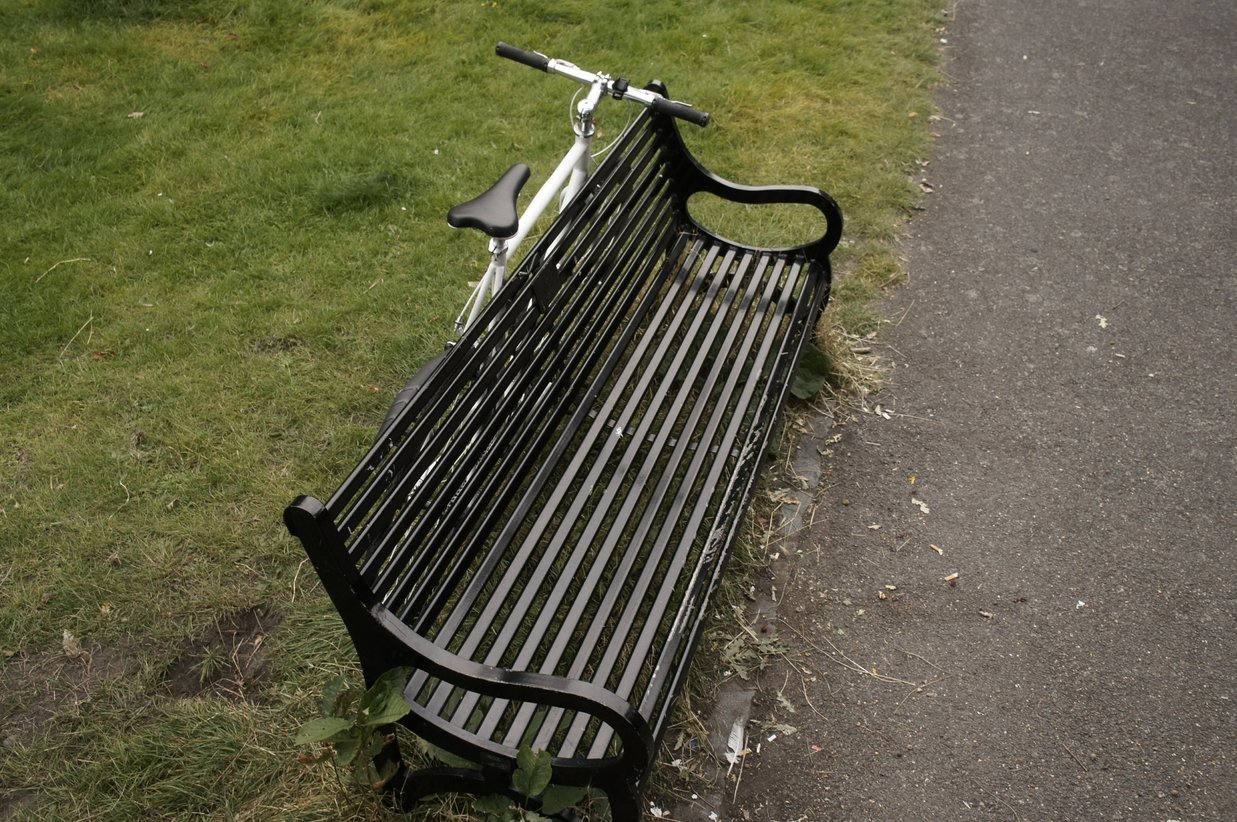} \\
$\lambda{=}0$ (no depth prior) & & Ground Truth \\
\end{tabular}

\caption{Ablation on $\lambda$ and $\tau$ for \textit{Bicycle}. Depth supervision improves RMSE but reduces RGB rendering quality (road texture), showing a geometry--appearance tradeoff under strong multi-view coverage.}
\label{fig:ablation_bicycle}
\end{figure*}

\Rhi{gpqz} {\subsection{Photometric Error as a Reliability Proxy}}

\Rhi{gpqz} {Table~\ref{tab:mask-validation} shows MAE is 1.16--1.19$\times$ lower inside the mask (4.14 vs.\ 4.82\,m at $\tau{=}0.18$, 95.9\% coverage; Table~\ref{tab:mask-significance}). Pixel-level differences are highly significant ($p{<}10^{-50}$); the frame-level paired $t$-test gives $p{=}0.047$ ($n{=}8$), confirming photometric error as a reliable proxy for depth quality.}

\paragraph{\Rhi{fxQT}{Comparison to a Depth-Inconsistency Mask.}}
\Rhi{fxQT}{Against a static, one-shot approximation of the Depth-Inconsistency Mask (DIM) of~\cite{in_depth_we_trust} at a matched reliable-pixel fraction (Table~\ref{tab:dim-comparison}), our mask separates accurate from inaccurate DA-V2 depth (ratio $1.17$) while the DIM proxy does not ($0.87$). This is expected: DIM keys on distortions that depth-supervised training itself induces, so a one-shot proxy has nothing to key on. The comparison is still the relevant one here, since our mask is likewise computed once and held fixed.}

\begin{table}[t]
\centering
\small
\resizebox{.95\linewidth}{!}{
\begin{tabular}{c c c c}
\hline
$\tau$ & MAE in / out (m) & AbsRel in / out & ratio \\
\hline
0.14 & 4.13 / 4.77 & 0.287 / 0.315 & 1.16 \\
\textbf{0.18} & \textbf{4.14} / \textbf{4.82} & \textbf{0.288} / \textbf{0.320} & \textbf{1.17} \\
0.22 & 4.15 / 4.94 & 0.288 / 0.326 & 1.19 \\
\hline
\end{tabular}
}
\caption {\Rhi{gpqz} {{Aligned DA-V2 depth error vs.\ GT LiDAR, inside vs.\ outside the photometric reliability mask, on 8 held-out RGB-only (KITTI Seq02, sparse every-2). 717{,}030 valid LiDAR pixels total.}}}
\label{tab:mask-validation}
\end{table}

\begin{table}[t]
\centering
\small
\resizebox{.95\linewidth}{!}{
\begin{tabular}{c c c c c}
\hline
$\tau$ & Mask \% & $n$ Inside & $n$ Outside & Cohen's $d$ \\
\hline
0.14 & 93.5\% & 670,745 & 46,285 & 0.13 \\
\textbf{0.18} & \textbf{95.9\%} & \textbf{687,417} & \textbf{29,613} & \textbf{0.14} \\
0.22 & 97.2\% & 697,183 & 19,847 & 0.16 \\
\hline
\end{tabular}
}
\vspace{6pt}

\resizebox{\linewidth}{!}{
\begin{tabular}{c c c c c}
\hline
$\tau$ & Mann-Whitney $p$ & Welch $p$ & Wilcoxon $p$ (Frame) & Paired $t$ $p$ (Frame) \\
\hline
0.14 & $3.6\times10^{-51}$ & $1.1\times10^{-100}$ & 0.039 & 0.048 \\
\textbf{0.18} & \textbf{5.5}$\times\mathbf{10^{-75}}$ & \textbf{7.4}$\times\mathbf{10^{-83}}$ & \textbf{0.074} & \textbf{0.047} \\
0.22 & $4.4\times10^{-100}$ & $7.9\times10^{-72}$ & 0.074 & 0.043 \\
\hline
\end{tabular}
}
\caption {\Rhi{gpqz} {Top: Mask partitions and effect sizes ($d$). Bottom: Pixel-level tests ($n\approx700$K) versus frame-level paired tests ($n=8$).}}
\label{tab:mask-significance}
\end{table}

\begin{table}[t]
\centering\small
\resizebox{\linewidth}{!}{%
\begin{tabular}{l c c c c c}
\toprule
Mask & Thresh & Frac. & \shortstack{MAE (m)\\in\,/\,out} & Ratio & $p_\mathrm{frm}$ \\
\midrule
Photometric (ours)                  & $\tau{=}0.18$        & 95.9\% & 4.14\,/\,4.82 & \textbf{1.17} & 0.047 \\
DIM proxy~\cite{in_depth_we_trust}  & $\varepsilon{=}17.6$m & 96.1\% & 4.19\,/\,3.66 & \textbf{0.87} & 0.995 \\
\bottomrule
\end{tabular}}
\caption{\Rhi{fxQT}{Photometric mask vs.\ static DIM proxy at matched pixel fraction. Ratio $=\mathrm{MAE}_\text{out}/\mathrm{MAE}_\text{in}$: ${>}1$ means the mask isolates accurate DA-V2 depth; ${<}1$ means no separation. AbsRel and Cohen's $d$ are in Tables~\ref{tab:mask-validation}--\ref{tab:mask-significance}.}}
\label{tab:dim-comparison}
\end{table}

\subsection{Object-Centric Bicycle Results}

\noindent\textbf{Takeaway.} In the object-centric \textit{Bicycle} scene~\cite{mipnerf360,nerfstudio}, depth supervision improves geometry as measured by RMSE, but it can hurt RGB rendering quality when multi-view coverage is already strong. \Rhi{fxQT}{Masking brings no gain over global supervision here, so the masked-vs-global advantage seen on KITTI does not transfer.}

Table~\ref{tab:bicycle_splatfacto_results} summarizes the hyperparameter sweep. The RGB-only baseline achieves the best rendering (17.731 PSNR, 0.570 SSIM, 0.240 LPIPS); depth supervision improves RMSE from 1.479 to 0.722 at the cost of rendering quality. \Rhi{fxQT}{Global supervision gives the best PSNR among depth-supervised runs (17.593 vs.\ 17.466 for the best mask), confirming the mask's rendering gain is specific to sparse forward-facing trajectories.}

\begin{table}[t]
\centering
\small
\resizebox{\linewidth}{!}{
\begin{tabular}{c c c c c c}
\hline
$\tau$ & $\lambda$ & PSNR$\uparrow$ & SSIM$\uparrow$ & LPIPS$\downarrow$ & RMSE$\downarrow$ \\
\hline
RGB-only & 0 & \textbf{17.731} & \textbf{0.570} & \textbf{0.240} & 1.479 \\
0.14 & 0.05 & 17.415 & 0.524 & 0.277 & 0.830 \\
0.14 & 0.15 & 17.036 & 0.477 & 0.309 & \textbf{0.722} \\
0.16 & 0.05 & 17.415 & 0.521 & 0.278 & 0.775 \\
0.18 & 0.05 & 17.466 & 0.520 & 0.279 & 0.751 \\
0.20 & 0.05 & 17.345 & 0.519 & 0.278 & 0.778 \\
0.22 & 0.10 & 17.106 & 0.491 & 0.297 & 0.725 \\
1.00 \Rhi{fxQT}{(global)} & 0.05 & 17.593 & 0.520 & 0.275 & 0.763 \\
\hline
\end{tabular}
}
\caption{
Compact summary of Splatfacto results on the \textit{Bicycle} scene. Depth supervision improves RMSE but generally reduces RGB rendering quality, showing a tradeoff between geometry and photometric fidelity in this object-centric setting.
}
\label{tab:bicycle_splatfacto_results}
\end{table}

\subsection{Discussion and Failure Modes}
The \Rhi{fxQT}{KITTISeq05 and KITTISeq00} results confirm the same trend: Splatfacto benefits from masked depth supervision while Mip-NeRF-360 degrades under both global and masked supervision. \Rhi{gpqz} {Unlike Splatfacto's direct per-Gaussian supervision, Mip-NeRF-360 distributes depth signals along rays, making implicit representations more sensitive to noisy priors.} \Rhi{fxQT}{Masking adds 0.44--0.70\,dB PSNR on KITTI at tied or better RMSE but yields no gain for Mip-NeRF-360 or \textit{Bicycle}: its benefit is rendering fidelity in explicit representations under weak multi-view coverage.}

\Rhi{gpqz} {Our numbers are based on DA-V2, but the explicit-vs-implicit architectural insights are prior-agnostic. The sweep values ($\tau$, $\lambda$) are dataset-specific. Our sweeps provide a tuning framework for new scenes and motivate future work on uncertainty-aware masks and adaptive depth-loss weighting.}

\section{Conclusion}
\label{sec:conclusion}

In this work, we investigate masked monocular depth supervision for sparse-view neural reconstruction using DA-V2 priors with Mip-NeRF-360 and Splatfacto on \Rhi{fxQT}{KITTI sequences 00/02/05} and the object-centric \textit{Bicycle} scene~\cite{kitti,mipnerf360,nerfstudio}.

The benefit depends strongly on backbone and scene. Splatfacto improves KITTISeq02 PSNR from 14.903 to 15.932 and RMSE from 0.542 to 0.100; \Rhi{fxQT}{against global (unmasked) supervision, the mask adds 0.44--0.70\,dB PSNR across KITTI 00/02/05 at tied or better RMSE, with no gain for Mip-NeRF-360 or \textit{Bicycle}: the benefit is rendering fidelity, not metric geometry.} Matched-ratio ablations confirm that gains come from selecting reliable low-error regions. \Rhi{fxQT}{The mask also outperforms three LiDAR-supervised baselines without ground-truth depth.}

Overall, monocular depth priors are most useful with explicit Gaussian representations, but less reliable for implicit NeRF-style representations.

{
    \small
    \bibliographystyle{abbrvnat}
    \bibliography{main}
}
\end{document}